\documentclass[a4paper,12pt]{article}

\usepackage[T1]{fontenc}
\usepackage{times}
\usepackage[english]{babel}
\usepackage[utf8]{inputenc}
\usepackage{dtklogos}
\usepackage{wallpaper}
\usepackage[absolute]{textpos}
\usepackage[top=2cm, bottom=2.5cm, left=3cm, right=3cm]{geometry}
\usepackage{appendix}
\usepackage[nottoc]{tocbibind}
\usepackage[colorlinks=true,
            linkcolor=black,
            urlcolor=blue,
            citecolor=black]{hyperref}
\usepackage{booktabs}

\usepackage{amsmath}
\usepackage{tabularx}
\usepackage{array}
\usepackage{ragged2e}
\usepackage{longtable}
\usepackage{sectsty}
\sectionfont{\fontsize{14}{15}\selectfont}
\subsectionfont{\fontsize{12}{15}\selectfont}
\subsubsectionfont{\fontsize{12}{15}\selectfont}

\usepackage{csquotes} 

\newsavebox{\mybox}
\newlength{\mydepth}
\newlength{\myheight}

\newenvironment{sidebar}%
{\begin{lrbox}{\mybox}\begin{minipage}{\textwidth}}%
{\end{minipage}\end{lrbox}%
 \settodepth{\mydepth}{\usebox{\mybox}}%
 \settoheight{\myheight}{\usebox{\mybox}}%
 \addtolength{\myheight}{\mydepth}%
 \noindent\makebox[0pt]{\hspace{-20pt}\rule[-\mydepth]{1pt}{\myheight}}%
 \usebox{\mybox}}

\newcommand\BackgroundPic{
    \put(-2,-3){
    \includegraphics[keepaspectratio,scale=0.3]{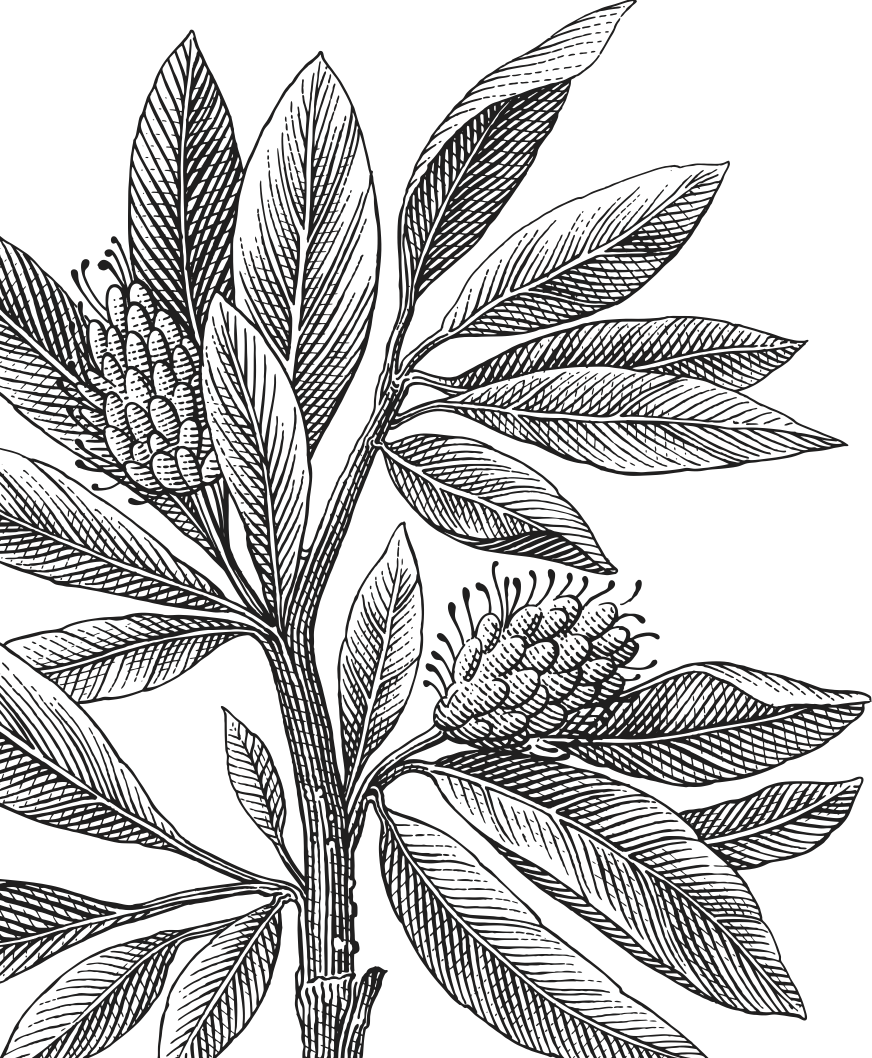} 
    }
}
\newcommand\BackgroundPicLogo{
    \put(30,740){
    \includegraphics[keepaspectratio,scale=0.10]{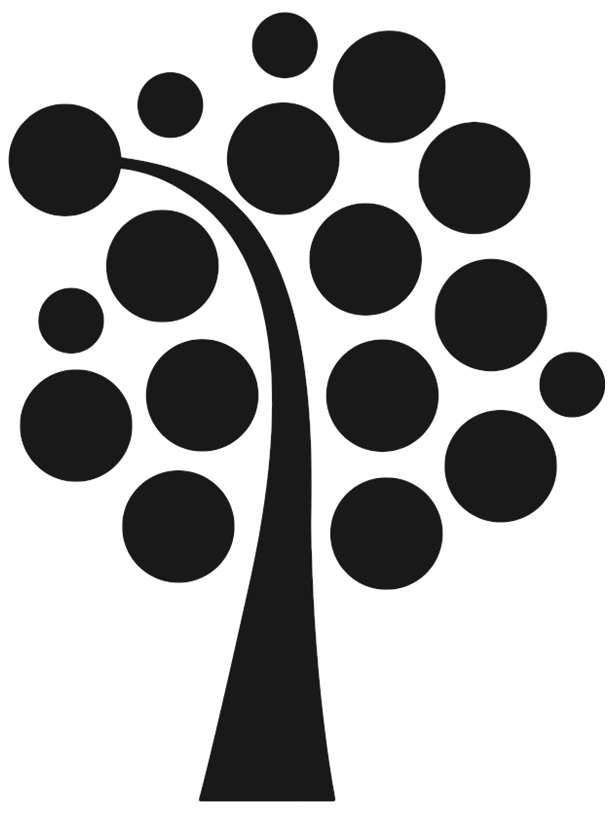} 
    }
}

\title{	
\vspace{-8cm}
\begin{sidebar}
    \vspace{10cm}
    \normalfont \normalsize
    \Huge Bachelor Degree Project \\
    \vspace{-1.3cm}
\end{sidebar}
\vspace{3cm}
\begin{flushleft}
    \huge Towards Interpretable and Efficient Feature Selection in Trajectory Datasets: A Taxonomic Approach \\ 

\end{flushleft}
\null
\vfill
\begin{textblock}{5}(10,12)
\begin{flushright}
\begin{minipage}{\textwidth}
\begin{flushleft} \large
\emph{Authors:} Chanuka Don Samarasinghage and Dhruv Gulabani\\ 
\emph{Supervisor:} Amilcar Soares\\ 
\emph{Examiner:} Rafael Messias Martins\\
\emph{Semester:} VT/HT 2025\\ %
\emph{Course:} 2DV50E \\ %
\emph{Subject:} Computer Science \\ 
\end{flushleft}
\end{minipage}
\end{flushright}
\end{textblock}
}

\date{} 

\begin{document}
\pagenumbering{gobble}
\newgeometry{left=5cm}
\AddToShipoutPicture*{\BackgroundPic}
\AddToShipoutPicture*{\BackgroundPicLogo}
\maketitle
\restoregeometry
\clearpage
\selectlanguage{english}
\begin{abstract}
\noindent Trajectory analysis is not only about obtaining movement data, but it is also of paramount importance in understanding the pattern in which an object moves through space and time, as well as in predicting its next move. Due to the significant interest in the area, data collection has improved substantially, resulting in a large number of features becoming available for training and predicting models. However, this introduces a high-dimensionality-induced feature explosion problem, which reduces the efficiency and interpretability of the data, thereby reducing the accuracy of machine learning models. To overcome this issue, feature selection has become one of the most prevalent tools. Thus, the objective of this paper was to introduce a taxonomy-based feature selection method that categorizes features based on their internal structure. This approach classifies the data into geometric and kinematic features, further categorizing them into curvature, indentation, speed, and acceleration. The comparative analysis indicated that a taxonomy-based approach consistently achieved comparable or superior predictive performance.
Furthermore, due to the taxonomic grouping, which reduces combinatorial space, the time taken to select features was drastically reduced. The taxonomy was also used to gain insights into what feature sets each dataset was more sensitive to. Overall, this study provides robust evidence that a taxonomy-based feature selection method can add a layer of interpretability, reduce dimensionality and computational complexity, and contribute to high-level decision-making. It serves as a step toward providing a methodological framework for researchers and practitioners dealing with trajectory datasets and contributing to the broader field of explainable artificial intelligence.



\end{abstract}

\newpage

\pagenumbering{gobble}
\tableofcontents 
\newpage
\pagenumbering{arabic}

%
%

\section{Introduction}

Trajectory data analysis is important to understand the movement patterns of moving objects and entities that include but are not limited to animals, vehicles, and natural phenomena. 
There is an unprecedented growth in trajectory data generated due to advancements in GPS, telecommunications, and remote sensing technologies \cite {rintoul2015trajectory,haidri2022ptrail,tavakoli2025taxonomical}.
This abundance of data opens up new opportunities for data mining and insights. 
It also poses significant challenges, particularly because of the high dimensionality of features extracted from these datasets; this problem is more commonly known as feature explosion  \cite{rintoul2015trajectory,tavakoli2025taxonomical,sadeghian2025review}.
Feature explosion complicates data analysis by increasing computing demands and costs, introduces redundancies, and reduces the interpretability and prediction accuracy of models \cite{rintoul2015trajectory,tavakoli2025taxonomical,sadeghian2025review}.

Feature selection is a process used to mitigate these issues. 
Basically, the subset of features that contribute most to the performance of a machine learning model is identified, thus improving computational efficiency, model accuracy, and interpretability \cite{singh2021feature}. 
Traditional feature selection methods can reduce the effect of some of these problems, but may not fully capture the complex hierarchical relationships between trajectory features \cite{tavakoli2025taxonomical}.

This work aims to investigate a taxonomy-based feature selection approach as a novel solution to these challenges in the context of trajectory datasets. 
Trajectory features are organized into meaningful categories based on their geometric and kinematic nature. 
This approach facilitates the reduction of dimensions more effectively while maintaining interpretability and improving the model's performance \cite{tavakoli2025taxonomical}.

This Bachelor's thesis in Computer Science, valued at 15 higher education credits (HEC), contributes to the fields of data mining and machine learning by developing, implementing, and evaluating a novel approach to the taxonomy-based feature selection method on real-world trajectory datasets. The study addresses the practical need for scalable, interpretable, and efficient methods to handle high-dimensional spatiotemporal data.

This chapter is organized as follows: Section 1.1 provides background and motivation for trajectory analysis and feature selection. Section 1.2 defines the research problem and questions. Section 1.3 outlines the objectives and contributions. Section 1.4  contains scope and limitations of this thesis. Section 1.5 defines our target audiences. Finally, Section 1.6 presents the thesis structure.

\subsection{Background}

A foundation is laid for analyzing movement patterns across diverse domains through trajectory datasets. 
These datasets are a collection of sequential records that are composed of ordered triplets - longitude, latitude, and timestamp values. 
A trajectory is formed when these values are combined in a time-ordered series, thereby tracing the path of an entity through space and time \cite{rintoul2015trajectory,haidri2022ptrail}. 
By examining these trajectories, researchers extract a wide range of movement characteristics such as speed, acceleration, indentation, and path curvature \cite{rintoul2015trajectory}. 
These extracted features uncover complex behavioral patterns, helping predict future movements and inform decision-making processes \cite{ribeiro2020survey}. 
With the advancement of data collection using machine learning techniques, including deep learning applications \cite{song2024enhancing,spadon2024multi,ferreira2022semi}, there has been a growing emphasis on utilizing these rich feature sets collected by mining spatiotemporal data to build predictive models capable of classifying trajectories by analyzing movement patterns \cite{tavakoli2022study,ferreira2023assessing}, building maritime networks \cite{carlini2020uncovering,carlini2022understanding}, detecting anomalies \cite{may2020challenges,abreu2021trajectory}, that are to support the development of novel approaches for trajectory analysis \cite{haranwala2022dashboard,haidri2022ptrail,li2021vessel,zhao2022deep}.

However, as the movement analysis has become sophisticated, so has the complexity of the underlying trajectory data grown, thereby increasing the number of features extracted from these trajectory datasets. 
By enabling the extraction of increasingly detailed and varied features from trajectory datasets, researchers now face the challenge of the proliferation of features commonly referred to as feature explosion  \cite{rintoul2015trajectory,tavakoli2025taxonomical}. 
The existence of numerous features may seem advantageous initially; however, this may also introduce some challenge known as feature explosion, which a well-known problem in machine learning where a large set of new features is extracted from a dataset, which results in the feature vector growing exponentially.

Datasets having a vast number of features are known as high-dimensional datasets. 
These datasets may complicate the analytical process in many ways. 
First, the dataset may have feature redundancy that can convey overlapping or similar information, thereby increasing the computational effort without providing any corresponding gain in machine learning model prediction. 
Second, feature redundancy can create ambiguity in extracting meaningful relationships from the data. So this makes it difficult for machine learning models for pattern recognition and reduces the overall performance.
This paradox is known as the “Curse of Dimensionality” \cite{sadeghian2025review,singh2021feature}. 
So, models are prone to overfitting and decreased predictive accuracy and efficiency when trained on these high-dimensional datasets. 
Due to high dimensionality, computational and storage requirements increases, which results in higher processing times and computational costs \cite{sadeghian2025review}.
 
In particular, feature selection, which aims to identify and retain only the most informative features from a potentially vast set, is crucial in addressing the issues associated with feature explosion. 
By appropriately handling high-dimensional data, feature selection not only improves model accuracy and computational efficiency but also enhances the interpretability of results, ultimately supporting better decision-making in trajectory analysis and related applications \cite{haidri2022ptrail,singh2021feature,li2021vessel,zhao2022deep}. 
These benefits are significant for trajectory analysis, where clear, explainable models are crucial for scientific and practical applications ranging from wildlife conservation and transportation optimization to developing natural disaster response plans \cite{rintoul2015trajectory,haidri2022ptrail,takeuchi2021frechet}.


In classical machine learning, feature selection methods belong to one of three categories: filter, wrapper, and embedded methods. 
Filter-based methods are known for their computational efficiency and scalability. 
In contrast, wrapper and embedded methods are known for their predictive performance at the cost of higher computational complexity \cite{sadeghian2025review,singh2021feature,borboudakis2019forward,kathirgamanathan2021feature}. 
Selecting a meaningful subset of features that improves model performance is a nontrivial task. This is because most standard feature selection techniques treat features as isolated variables, providing limited insight into why specific features were selected and reducing the interpretability of the model. This leaves room for research into feature selection methods that have a structured and explainable approach towards feature selection \cite{sadeghian2025review,singh2021feature,li2021vessel}.

In the context of trajectory analysis and movement data, the need for such a transparent feature selection process has become a large area of interest \cite{tavakoli2025taxonomical}. 
Based on this, recent work within the trajectory analysis successfully arranged these trajectory features into structured taxonomies and generated models with high interpretability and predictive performance \cite{tavakoli2022study,tavakoli2025taxonomical}. 
Building on these developments, this study proposes a taxonomy-based feature selection framework that organizes features into hierarchical groups based on geometric and kinematic properties. 
By aligning the feature selection process with a taxonomy that reflects the natural structure of movement data, this approach aims to provide both dimensionality reduction and improved explainability, key priorities in the emerging field of eXplainable AI (XAI) \cite{tavakoli2025taxonomical,guyon2003introduction,nakanishi2024evolving}.

\subsection{Research Problem, Questions and Motivation}

The rapid expansion of trajectory datasets has amplified the problem of feature explosion, where an exceptional increase in derived features complicates data analysis and diminishes the effectiveness of classical methods \cite{rintoul2015trajectory,haidri2022ptrail}. 
Although classical feature selection techniques help address high dimensionality, they typically treat features as independent variables and fail to encapsulate their underlying relationships \cite{sadeghian2025review,singh2021feature,borboudakis2019forward}. 
This often limits interpretability, thereby making it challenging to understand the rationale behind model decisions.
This thesis proposes a novel taxonomy-based feature selection approach to address feature explosion in trajectory datasets. 
The main research problem is to determine whether this method can reduce high dimensionality and improve efficiency, effectiveness, accuracy, and interpretability in comparison to other traditional feature selection techniques.
This work is motivated by the need for more efficient and interpretable analysis of large trajectory datasets in practical applications. 
It advances data mining and machine learning approaches by promoting more explainable models through suggested taxonomy-based feature selection. 
Based on these considerations, the following research questions guided this thesis:

\begin{enumerate}
    \item \textbf{(RQ1)} \textit{Investigate.} How do existing feature selection approaches perform in the context of high-dimensional trajectory data?
    
    \item \textbf{(RQ2)} \textit{Evaluate.} How would a taxonomy-based feature selection approach, which systematically organizes features set into related categories, perform in the context of trajectory data?
    
    \item \textbf{(RQ3)} \textit{Evaluate.} How does the proposed taxonomy-based method compare to conventional feature selection techniques in terms of computational efficiency, performance, and model explainability?
\end{enumerate}

Our underlying hypothesis was that the suggested taxonomy-based approach will achieve at least comparable, if not superior, predictive performance while providing a higher level of interpretability than standard and traditional methods. 
In pursuing answers to these questions, this thesis offers both a methodological advancement and practical guidance for researchers and practitioners working with complex trajectory data.

\subsection{Objectives and Contributions}

The main objective of this thesis is to develop, analyze, and evaluate a taxonomy-based feature selection framework for trajectory analysis. 
The work is structured around three specific aims: first, to conduct an analysis on existing feature selection techniques for high-dimensional datasets; 
second, to carefully implement a taxonomy-based approach that systematically groups features according to geometric and kinematic properties based on Yashsar et al. {\cite{tavakoli2025taxonomical}}'s  work
and third, to empirically compare the proposed framework with classical feature selection methods in terms of classification accuracy, computational efficiency, and model interpretability.
The primary scientific contribution of this study is the introduction of a structured taxonomy for organizing trajectory features into standard feature selection methods, which is expected to facilitate more interpretable and efficient dimensionality reduction. 
Through a series of controlled experiments on real-world movement datasets, this research provides empirical evidence on the relative merits of taxonomy-based feature selection in practical machine learning tasks. 
In addition to methodological advancement, the thesis offers a reproducible pipeline, including all scripts and algorithms, enabling other researchers and practitioners to apply and extend the proposed framework in diverse domains where movement data analysis is required. 

This study's outcomes are expected to benefit the scientific community and have broad practical applications that contribute to society at large. 
They can range from supporting wildlife conservation through analyzing migration patterns to analyzing vessel traffic, designing ports or other transportation infrastructure, and even improving urban mobility planning or disaster response. 
The proposed method enables more informed, data-driven decisions in these and other application areas by providing higher levels of interpretability in the context of trajectory data.

\subsection{Scope and Limitations}

The scope of this thesis includes the investigation, development, application, and evaluation of the taxonomy-based feature selection approach, with experiments conducted on movement data from diverse backgrounds, including wildlife, hurricanes, and maritime vessels. 
The results and conclusions are specifically validated on the features selected from these datasets, although the methodology is designed to be broadly applicable to other trajectory datasets. 
The results of this study are the limited range of datasets analyzed, possible computational and timing constraints, and necessary simplifications in taxonomy construction. 
Furthermore, four machine learning models were selected for the analysis. 
These factors may affect the generalization of the results. 

\subsection{Target Audience}
\noindent Following is the identified target audience for the study :
\begin{itemize}
    \item Researchers in the broader machine learning domain.
     \item Researchers interested in any trajectory-based phenomena
    \item Designers and architects depend on trajectory data to get insights into their designs.
\end{itemize}

\subsection{Thesis Structure}
\noindent Following is the structure of this thesis.
\begin{itemize}
    \item \textbf{Chapter 2 - Related Work:} Summarizes previous research and existing solutions relevant to the project.
    \item \textbf{Chapter 3 - Methodology:} Describes the methods and techniques used to conduct the study.    
    \item \textbf{Chapter 4 - Results and Analysis:} Presents and analyzes the findings of the study.
    \item \textbf{Chapter 5 - Discussion:} Interprets the results, discussing their implications, limitations, and possible improvements.
    \item \textbf{Chapter 6 - Conclusion and Future Work:} Summarizes the study's contributions and suggests directions for future research.
\end{itemize}

\newpage

\section{Related Work}

The following chapter analyses the current research on the topic of feature selection and taxonomy-based feature categorization.  Next, a comparative analysis of the current methods and their characteristics is discussed. Finally, based on the comparative analysis, the theoretical gaps can be identified and then used to justify and position this study in the broader context of machine learning and feature selection.

\subsection{Review of existing research regarding to Feature Selection}

Kathirgamanathan and Cunningham {\cite{kathirgamanathan2021feature}} introduces a new feature selection method in their work, which calculates a merit score to select the best subset of features. The merit score is calculated by first training individual models for specific time series features and only those features.  Afterwards, the correlation between each feature classifier’s output and the class labels is collected, which they define as feature-to-class correlation. Additionally, the correlation outputs between the different single-feature classifiers are calculated, which they define as feature-to-feature correlation.  Then they use the following formula to calculate the merit score metric.

\[
    \text{Merit}_S = \frac{k \cdot \bar{r}_{cf}}{\sqrt{k + k(k - 1)\cdot \bar{r}_{ff}}}
\]

where:
\begin{itemize}
  \item $k$ is the number of features in the subset,
  \item $\bar{r}_{cf}$ is the average correlation between each feature and the class,
  \item $\bar{r}_{ff}$ is the average correlation between the features.
\end{itemize}

The formula is designed to favor feature subsets that are highly correlated with the classes and uncorrelated with each other. Although this study considers the features' correlation, it fails to do so in the context of their structure. Furthermore, using these correlation calculations to select the features fails to add a layer of interpretation to improve the explainability of the results on a domain level. Other than that, this method needs to consider each feature separately, where the computation time is increased compared to a taxonomy-based method, where the feature categories are used as combinations \cite{kathirgamanathan2021feature}. 

Sadeghian et al. {\cite{sadeghian2025review}} review meta-heuristic algorithms that evaluate solutions through Swarm Intelligence while performing space searches in parallel. They are guided to use only one fitness function to search complex solution spaces for performing feature selection, which results in faster feature selection. Compared to classical approaches, Meta-Heuristic Algorithms perform better on high-dimensional datasets due to low computational costs and high levels of flexibility, as they have low sensitivity for initial values [4]. However, these methods fail to account for the internal structural relationships among features, and as a result, these complex algorithms negatively impact the interpretability of their output compared to a taxonomy-based method. 	

Gao et al.  {\cite{gao2023multi}} propose a feature selection method based on clustering and information entropy. In their approach, they use k-means to cluster all the features based on their similarity. Then, using information entropy, the informativeness of each feature on its ability to differentiate between different classes is calculated. Afterwards, the features with high informativeness are retained in each cluster while the rest are discarded.  Finally, using the retained clusters, a new feature subset is created, where the subset is expected to have low levels of redundancy and high relevance \cite{gao2023multi}.  This study employs a clustering technique to group similar features; however, the use of information entropy does not account for the internal structure of the features within the domain of the problem, making it more challenging to interpret the selection results compared to a taxonomy-based method. 

In their paper, Nakanishi et al. {\cite{nakanishi2024evolving}}propose a feature selection method called Synergistic forward and backward selection. They use “Approximate Inverse Model Explanation” (AIME) to calculate a global feature importance measure to organize the features from best to worst. Then, a backwards feature deletion process is carried out where the feature with the least score is eliminated, and the model is retrained. This process is iteratively carried out until the smallest subset is selected while retaining the model’s maximum accuracy.  A forward deletion step is carried out in the next step, where features are removed individually to check if the model maintains accuracy. If so, that feature is removed, and this is iterated until the smallest possible subset remains while retaining the model’s accuracy \cite{nakanishi2024evolving}. Even though the authors argue this adds a level of interpretation to the data, it does so on an individual feature level based on an AIME score. This does not translate into domain-specific high-level information that the decision makers can use to make informed decisions.   

Liu et al. {\cite{liu2019cost}} propose a feature selection that uses F-measure instead of the commonly used accuracy metric to select the features. First, they assign an importance score for each feature based on its ability to classify all classes individually based on the F-measure. Using this, a cost vector is created for each F-measure, where a higher score is given if a minority class is misclassified. Next, using each cost vector, a cost-sensitive regularization
problem is solved, which produces one model for the individual F-measure. Then, using the validation set, the models are re-evaluated with F-measure and the best model is selected \cite {liu2019cost}. However, due to its complex feature selection process, this method offers little interpretability compared to a taxonomy-based method.

\begin{figure}[h]
    \centering
    \includegraphics[width=1.0\textwidth]{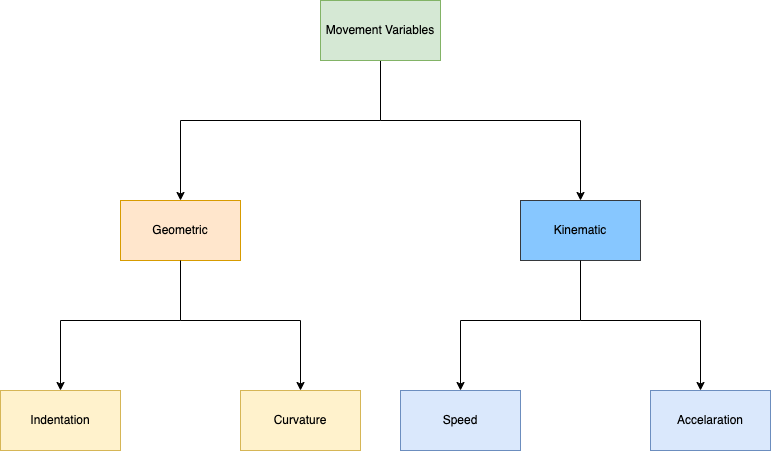}
    \caption{Taxonomy}
    \label{fig:Taxonomy}
\end{figure}

The work Yashar et al. {\cite{tavakoli2025taxonomical}} has introduced a taxonomy-based approach to reduce the dimensionality of movement data and improve the explainability of movement data. 
Their work uses a taxonomy that splits the trajectory features based on geometric and kinematic features.  Geometric features represent the shape of the trajectory, whilst kinematic features represent the behavior encompassing the significant components of any given trajectory. 
Then, at a lower level, geometric features contains indentation and curvature-based features while kinematic features  contains speed and acceleration-based features \cite{tavakoli2025taxonomical}. 
The specific features within these categories are available in Table \ref{tab:movement_variables} while Figure \ref{fig:Taxonomy} visually represents the proposed taxonomy.

\begin{table}[h]
\centering
\caption{Description of movement-related variables and categories~\cite{tavakoli2025taxonomical}}
\label{tab:movement_variables}
\begin{tabular}{|p{3cm}|p{4cm}|p{7cm}|}
\hline
\textbf{Leaf} & \textbf{Movement Variables} & \textbf{Description} \\
\hline
Curvature & 5 signatures of Distance Geometries consisting of 15 variables & 15 variables that measure straightness through effective distance (the ratio of the distance between the start and end points of a segment to the length of the segment, where 1 indicates a straight trajectory and a value close to 0 indicates a tortuous one). Each summation term (signature) measures straightness in progressively finer frequencies. \\
\hline
Indentation & 19 statistical variables: unique values, number of 0’s, mean, standard error, quantiles, std. dev., coefficient of variation, MAD, IQR, skewness, kurtosis, min, max & These summarize the distribution of angles (indentation) at each point visited by the moving object across its trajectory. \\
\hline
Speed & 19 statistical variables (same as Indentation) & Summarizes the distribution of speed magnitude at each point over the movement. \\
\hline
Acceleration & 19 statistical variables (same as Indentation) & Summarizes the distribution of acceleration magnitude at each point over the movement. \\
\hline
\end{tabular}
\end{table}

Overall, they successfully identified multiple interesting movement patterns and accordingly, which taxonomy components contributed the most in explaining the movement patterns \cite{tavakoli2025taxonomical}. However, this study is mainly aimed at classifying unlabeled data using clustering methods which is not in the context of feature selection.

\newpage

\subsection{Comparison of existing research}
Various methodologies have been proposed to advance feature selection for trajectory and movement data analysis. Spanning from the classical filter, wrapper, and embedded methods to modern information-theoretic, clustering-based and meta-heuristic approaches and an easily interpretable taxonomy-based approach. Each of these methods presents unique strengths and specific limitations. These depend on factors such as the model interpretability requirements, dimensionality, data structure, and inherent complexity of trajectory patterns analyzed. The Table {\ref{tab:comparison_table}} provides a critical comparison of diverse approaches. It highlights their methodological focus towards effectiveness, efficiency, scalability, interpretability and suitability, along with strengths and limitations. This analysis contextualizes this study in the feature selection field as this thesis's goal.

\vspace{1em}

\begin{longtable}{|
  >{\RaggedRight\arraybackslash}p{2.9cm}|
  >{\RaggedRight\arraybackslash}p{3.6cm}|
  >{\RaggedRight\arraybackslash}p{3.8cm}|
  >{\RaggedRight\arraybackslash}p{3.8cm}|
}
\caption{Comparison of Existing Research}
\label{tab:comparison_table} \\
\hline
\textbf{Study / Reference} & \textbf{Approach / Domain} & \textbf{Strengths} & \textbf{Limitations} \\
\hline
\endfirsthead
\hline
\textbf{Study / Reference} & \textbf{Approach / Domain} & \textbf{Strengths} & \textbf{Limitations} \\
\hline
\endhead
Kathirgamanathan \& Cunningham (2021) & Unsupervised feature selection for time-series data in high-dimensional, unlabeled scenarios & Effectively identifies relevant features in unlabeled, high-dimensional datasets; robust merit score for feature selection & Lacks interpretability at the domain level; complex computations increase processing time \\
\hline
Sadeghian et al. (2023) & Meta-heuristic algorithms for feature selection across a wide range of high-dimensional real-world domains & Comprehensive, up-to-date review; adaptable to various data types; improves performance in high-dimensional settings & No universal best algorithm; results sensitive to dataset and parameter tuning; interpretability and explainability are not emphasized \\
\hline
Wu Xia et al. (2021) & Feature selection and combination for aerospace target recognition using K-means clustering and information entropy & Improves classification accuracy; reduces computation time; effective for high-dimensional sensor data; integrates clustering with information theory & Primarily validated for aerospace data; less sensitive to subtle features; lacks interpretability; not designed for trajectory or time-series data \\
\hline
Nakanishi et al. (2024) & Explainable artificial intelligence and multimodal feature selection in high-dimensional data using global feature importance (AIME) and backward-forward deletion & Provides global and local feature interpretability; model-agnostic; enables systematic elimination of less relevant features; transparent and efficient for large datasets & Performance gains limited if feature importances are similar; can be computationally intensive for large datasets; adaptation needed for trajectory or time-series data \\
\hline
Liu et al. (2019) & Cost-sensitive feature selection for imbalanced multi-class and multi-label machine learning using F-measure optimization & Addresses class imbalance by optimizing F-measure; selects features for both majority and minority classes; theoretical guarantees; efficient optimization; empirically strong performance & Optimizes only for F-measure; interpretability not a priority; mainly for tabular or image data—not trajectory or movement data \\
\hline
Tavakoli, Peña-Castillo \& Soares (2025) & Taxonomy-driven analysis of high-dimensional, unlabeled movement data using multilevel taxonomical description and anomaly detection & Enables interpretable summaries of complex, unlabeled movement datasets; uncovers meaningful patterns via taxonomy grouping; reduces dimensionality while maintaining interpretability; validated on multiple real-world datasets & Depends on taxonomy quality and parameters; sensitive to outlier detection method; requires expert input for taxonomy creation; primarily descriptive—not a direct feature selection method for predictive modeling \\
\hline
\end{longtable}

\subsection{Positioning of this work}

Addressing high dimensionality, redundancy, and computational complexity is the core
objective of feature selection in general.  Based on the related work, it is evident that there is no one-size-fits-all solution available for this problem, and there is room for more novel methods for feature selection. Most current studies lack interpretability or have high computational cost due to the possible subsets available for feature selection, which is a significant concern in the current context \cite{liu2019cost, sadeghian2025review}. Overall, the following gaps were identified.

\begin{enumerate}
    \item The level of interpretability could be improved, particularly to a level that can support data-driven decision-making.
    \item The number of feature subsets could be reduced systematically to speed up the feature selection process; however, using a taxonomy for this purpose has not been studied.
    \item Taxonomy-based methods are yet to be evaluated in the context of feature selection for labeled data.
\end{enumerate}

Based on this, the applicability of this approach in the context of labeled trajectory data and feature selection raises an interesting question that needs to be investigated. As such, this study focuses on the hypothesis that a taxonomic feature selection method in the context of labeled trajectory data can provide better results than classical methods.

\newpage 

\section{Methodology}
\label{Method}

The main objective of this study is to evaluate whether a taxonomy-based feature selection approach can outperform traditional feature selection methods in the case, specifically backward and forward selection, while adding a layer of interpretability to explain a model's behavior. 
A machine learning pipeline was developed to achieve this goal, and a set of controlled experiments was carried out, gathering a set of performance metrics for in-depth analysis. 
The pipeline consists of data collection, feature extraction, feature selection, feature scaling, model training, hyperparameter tuning, performance metric generation, and plotting the results. 
Through a comparison analysis of the generated plots, the study assesses the validity of the proposed taxonomy-based feature selection approach.

\subsection{Proposed machine learning pipeline}

\begin{figure}[h]
    \centering
    \includegraphics[width=1.1\textwidth]{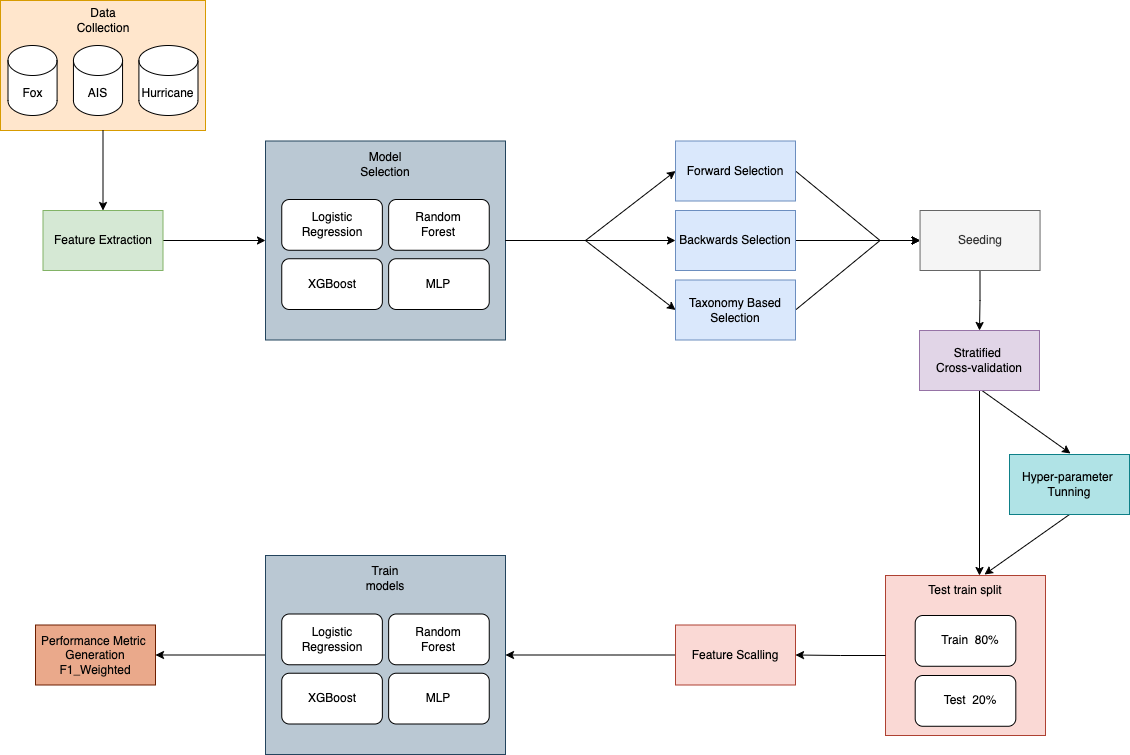}
    \caption{Pipeline overview}
    \label{fig:Pipeline overview}
\end{figure}

Figure \ref{fig:Pipeline overview} shows an overview of the proposed pipeline.
Our pipeline starts at the data set selection stage, where three data sets were selected for the analysis. 
Based on these datasets, features related to overall kinematic and geometric statistics are extracted. 
After the feature extraction, the feature selection is performed using three different approaches, namely forward, backward, and taxonomy-based feature selection. 
Then, the set of experimentation iterations is defined by employing four different seeds and five stratified cross-validations. 
Then, the experiment branches out to where the models are trained directly and where hyperparameter tuning is employed. 
Afterward, a feature scaling step brings all the selected features to a similar scale. 
Next, the models are trained, and finally, performance metrics are generated for analysis. 
Our study is carried out in a systematic manner to evaluate whether the proposed taxonomy-based approach yields better results compared to the traditional methods selected by keeping all the steps standard and unified for each feature selection method.

\clearpage

\subsection{Datasets}
Three distinct datasets were selected for the study, related to animal migration trajectories, ship trajectories, and hurricane movement trajectories, to test the approach on a wide application area, and are detailed below. \\Each of these datasets contained latitude, longitude time stamp, trajectory\_id and a classifier within them as base features. 


\noindent\textbf{Arctic Fox dataset.} This dataset contains a set of trajectories of arctic fox. 
The data set itself is small, and it is used to classify whether the trajectory belongs to a male or a female. 
As such, the total data set was used for the study. 
The Table \ref{tab:label_distribution} summarizes the sample sizes \cite{lai2016arcticfox}.

\begin{table}[h]
\centering
\caption{Sample distribution by label}
\label{tab:label_distribution}
\begin{tabular}{|c|c|}
\hline
\textbf{Label} & \textbf{Trajectory Samples}\\
\hline
F & 38\\
M & 28\\
\hline
\end{tabular}
\end{table} 

\noindent\textbf{AIS dataset.} This dataset contains the trajectories of different types of vessels in a port area and was extracted from \cite{vtexplorer2025ais}. 
This specific dataset was large and heavily imbalanced in its raw format. 
Due to time limitations and the imbalances in the data set, a decision was made to reduce the data set to a specific set of classes that was more balanced in comparison. This was done through analyzing the dataset and selecting the top four samples classes that had similar frequency compared to the others and had over 100 sample at least. The  Table {\ref{tab:vessel_classes}} summarizes the selected classes.

\begin{table}[h]
\centering
\caption{Summary of selected vessel classes and their sample counts}
\label{tab:vessel_classes}
\begin{tabular}{|c|l|c|}
\hline
\textbf{Label} & \textbf{Vessel Type} & \textbf{Samples} \\
\hline
30& Fishing & 250 \\
52& Tug & 171 \\
60& Passenger, all ships of this type & 161 \\
70& Cargo, all ships of this type & 112 \\
\hline
\end{tabular}
\end{table}

\noindent\textbf{Hurricanes.} This dataset contains tropical cyclones' trajectories over a couple of centuries in total and where they originated. Thus, a more standardized approach was required to reduce the data set in order to manage the study within the given time limitations. 
As such, Scikit-Learn's resampling method was used to collect a sample set of one thousand from the top five classes available in the dataset, with the random state set to 42. 
The Table \ref{tab:regional_distribution} summarizes the sample classes \cite{knapp2018ibtracs}.   

\begin{table}[h]
\centering
\caption{Sample distribution across regional labels}
\label{tab:regional_distribution}
\begin{tabular}{|c|c|}
\hline
\textbf{Label} & \textbf{Samples} \\
\hline
WP & 200 \\
SI & 200 \\
NI & 200 \\
EP & 200 \\
SP & 200 \\
\hline
\end{tabular}
\end{table}

In the trajectory datasets the occurrence of stationary targets produces NaN type results in some of the features. However, some of the models does not accept such inputs.Thus, a data imputation using the SimpleImputer form Scikit-Learn was used fill the missing data. The Venn diagram at figure 3.3 visualizes the expected combinations.

\subsection{Feature extraction}

The feature extraction is based on the work of Yashar et al. \cite{tavakoli2025taxonomical}. 
As the first step, each trajectory was grouped by its unique trajectory identifier and was ordered chronologically in order to preserve the natural order of the movement. 
Afterwards, the specific features pertaining to curvature (distance geometry), indentation (angle), speed, and acceleration were calculated through the use of Scipy, Geopy, and Numpy libraries. 
Table \ref{tab:movement_variables} summarizes the extracted features for each type of feature as described by \cite{tavakoli2025taxonomical}.

\subsection{Model selection}

For the study, four different models were selected, covering four of the major classifier paradigms available in the context of machine learning. 
They are Logistic Regression, Random Forest, XGBoost, and Multilayer Perceptron. 
Logistic regression is a widely used linear classification model for binary and multiclass classification. At the model's core is a logistic (sigmoid if binary or a softmax if multiclass) function that is used to compute the probabilities. Despite the model's simplicity, it is well regarded for its interpretability and computational efficiency  \cite{shwartz2022tabular}.
Random forest is an Ensemble (Bagging) type of machine learning model that constructs many decision trees. It combines these to improve the predictive performance and reduce overfitting. It is known for its ability to work with complex and noise-filled datasets \cite{breiman2001random}.
XGboost is an Ensemble (Boosting) type of machine learning model. It builds ensembles of decision trees sequentially, in which each consecutive tree fixes the errors of the previous tree. It's known for its speed and predictive performance in large-scale datasets \cite{chen2016xgboost}.
Multilayer Perceptron (MLP) is a Feedforward Neural Network type of a machine learning model that contains layers of neurons capable of learning complex nonlinear relationships in the dataset \cite{rumelhart1986learning}. 
In the study, in order to implement these models, Scikit-Learn was used for all models except XGBoost, where the XGBClassifier library was used.

\subsection{Feature selection approaches}

Feature selection plays a significant role in improving a model's prediction accuracy, reducing dimensionality, and increasing the interpretability of the model \cite{kathirgamanathan2021feature, guyon2003introduction}. 
This study aims to introduce a novel feature selection approach based on taxonomy and compare the results; thus, a baseline must be initialized for comparison. 
Backward and forward feature selection is one of the most common techniques used in the machine learning environment. 
These greedy wrapper-based approaches add or remove features based on their contribution to the model's performance. 
However, both these approaches leave little room for interpretation of the feature selection process due to its black-box nature. 
As such, it was decided that these two would be valid candidates for initializing a baseline for this study. \\

\noindent \textbf{Baseline feature selection methods.} The forward selection procedure begins with an empty set of features and adds features iteratively. Then, the features that increase the model's the most are identified, and those with negative or minimal contributions are ignored. 
This process continues until the model's performance is no longer increasing or some predefined condition is met (IE: a predefined number of features are filled, which stops the algorithm) \cite{guyon2003introduction}.  “Sequential Feature Selector” from the Scikit-Learn library was used to implement this feature.
In contrast, to the forward selection, the backward selection starts with the whole set of features and iteratively removes the least significant feature. The stopping criterion is similar where the process goes on until a feature can no longer be removed without reducing the model's performance or a predefined condition is met \cite{guyon2003introduction}. “Sequential Feature Selector” from the Scikit-Learn library was also used here to implement this feature.\\ In this study, the selection process was stopped when performance is no longer increasing for both forward and backward selection.

\noindent \textbf{Taxonomy-based feature selection.}
Trajectory data in their raw form lacks comprehensiveness and interpretability due to its complex nature. 
However, one solution to alleviate this problem is to systematically organize the inherent properties of these movements into a taxonomy that helps interpretability through a hierarchical representation of its structure \cite{tavakoli2025taxonomical}.

Based on this approach, feature selection itself is done at the given taxonomic level, thereby abstracting individual features and taking all possible combinations at all levels, which adds a new level of interpretation to feature sets that are selected. The total combinations, excluding the null set, can be calculated using the simple formula $2^n - 1$ where n = 4 because the lowest level of the taxonomy contains four categories such as curvature (C), indentation (I), speed (S) and acceleration (Ac). Therefore, this taxonomy produces 15 different combinations of feature sets to train the models, which is visualized in Figure {\ref{fig:TaxonomyCombinations}. 

\begin{figure}[h]
    \centering
    \includegraphics[width=0.7\textwidth]{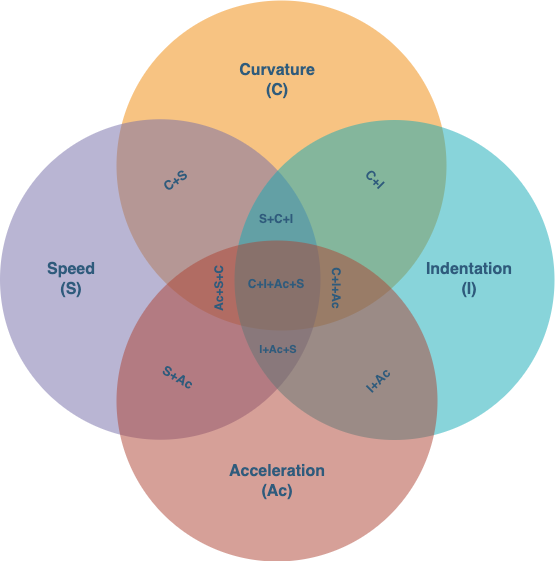}
    \caption{Taxonomy combinations}
    \label{fig:TaxonomyCombinations}
\end{figure}

\subsection{Seed Control and Stratified Cross-validation}

The experiment was carried out using different seeds, using five decimal places of $\pi$: 14159, 26535, 89793, 23846, respectively, as the seeds. 
Then, five stratified cross-validations were performed for each seed, resulting in five different test and train datasets and twenty different iterations of the experiment for each selected model. These steps were integrated to improve the reliability and validity of the study by ensuring that the results are not based on a random seed or a single split of the data set. 

Then, in the context of the baseline for each selected approach, backward and forward selection, this pipeline is run separately, resulting in twenty sets of individual features selected for each approach. 
Next, for the taxonomy approach, the same pipeline is used for each of its produced combinations, producing twenty sets of results for each combination.

\subsection{Hyperparameter Tuning Setup}

Hyperparameter tuning is the process of identifying the best set of hyperparameters for a given model that optimizes its performance. Introducing this step ensures that the evaluation is carried out in the best possible configuration, contributing to the reliability and validity of the study further. Different models have different hyperparameters, and for this specific study, Table \ref{tab:hyper_params} presents the selected hyperparameters and their values. Using these hyperparameters, a “Grid Search”  was carried out using the Scikit-Learn library to perform hyperparameter tuning for this study.

\begin{table}[ht]
\centering
\caption{Model hyperparameters, value ranges, and descriptions.}
\label{tab:hyper_params}
\resizebox{\textwidth}{!}{%
\renewcommand{\arraystretch}{1.6}
\begin{tabular}{|l|l|l|l|}
\hline
\textbf{Model} & \textbf{Hyperparameter} & \textbf{Values} & \textbf{Description} \\
\hline
Logistic Regression & c & 0.1, 1.0, 10.0 & Inverse of regularization strength \\
Logistic Regression & penalty & l1, l2 & Regularisation type \\
Logistic Regression & solver & liblinear, saga & Optimization algorithm \\
\hline
Random Forest & n\_estimators & 100, 500, 1000 & Number of trees in the forest \\
Random Forest & max\_depth & None, 10, 20 & Max depth of each tree (None = unlimited) \\
Random Forest & max\_features & sqrt, log2, 16 & Number of features considered for best split \\
\hline
XGBoost & n\_estimators & 100, 500, 1000 & Number of boosting rounds/trees \\
XGBoost & max\_depth & 3, 6, 10 & Max depth of each tree \\
XGBoost & learning\_rate & 0.01, 0.1 & Step size shrinkage (used to prevent overfitting) \\
XGBoost & sub\_sample & 0.8, 1.0 & Sample fraction used for fitting each tree \\
\hline
MLP & hidden\_layer\_sizes & (50,), (100,), (50, 50) & Number and size of hidden layers \\
MLP & alpha & 1e-4, 1e-3, 1e-2 & l2 regularization parameter \\
MLP & learning\_rate\_init & 1e-4, 1e-3, 1e-2 & Initial learning rate for weight updates \\
\hline
\end{tabular}
}
\end{table}

\subsection{Scaling and training of models}

Prior to fitting the models with the data using the selected features, a scaling step is included in order to equalize the contribution of each feature through Scikit-Learn’s “Standard Scalar.” Next, the models are trained on the selected feature sets for each iteration using the standard 80/20 data split. 

\subsection{Performance Metric Selection and Visualization of Results}

F$1$ measure is one of the most commonly used and effective metrics for evaluating model performance \cite{liu2019cost}. 
Three main variations of F$1$ scores are available: Macro-F$1$, Micro-F$1$, Weighted-F$1$. The following briefly introduces each variant and how it is calculated \cite{kathirgamanathan2021feature}.  

Macro-F1 calculates the F$1$ score for each class separately and the unweighted mean. This means it treats all classes equally, regardless of their frequency within the dataset. The calculation formula is given below. 

\[
\mathrm{Macro\ F1} = \frac{1}{N} \sum_{i=1}^{N} \mathrm{F1}_i
\]

Micro-F1 aggregates the contribution of each class to calculate the average F1 score, which is more affected by the results of the more frequent classes available in the dataset. It is calculated as below.

\[
\mathrm{Micro\ F1} = \frac{2 \cdot TP_{\mathrm{total}}}{2 \cdot TP_{\mathrm{total}} + FP_{\mathrm{total}} + FN_{\mathrm{total}}}
\]

Weighted-F1 calculates the F1 score for each class and takes an average. Then, it is weighted by the actual occurrence of that class. 
This results in a value that takes into consideration the minority classes and the contribution of each class as well. It is calculated as below.

\[
\mathrm{Weighted\ F1} = \frac{\sum_{i=1}^{N} \mathrm{support}_i \cdot \mathrm{F1}_i}{\sum_{i=1}^{N} \mathrm{support}_i}
\]

Existing research suggests that the Weighted-F1 score is a better predictor for time series datasets \cite{kathirgamanathan2021feature}. 
Thus, the Weighted-F1 score is selected as the performance metric to be used in this study. 
Finally, the gathered data is visualized through box plots individually for specific datasets and specific models.

\subsection{Limitations faced}

The study was developed in a controlled, systematic manner to conduct a robust analysis of the proposed hypothesis. 
However, the study faced limitations while running the experiment, mainly due to time constraints. 
First, the sample collection had to be limited to around a thousand to support the available computation capabilities. Next, hyperparameter testing rounds were limited in the context of hyperparameter selection due to the constraints above. 
Furthermore, the model selection could include more variety to check it in different contexts. 
Finally, the obtained results can be used to further optimize the taxonomy approach, refining it, and comparing it against other available feature selection methods.

\subsection{Ethical Considerations}

There are no essential ethical considerations because no human subjects were involved in this study, and this work does not deal with any specific sensitive topic in this direction. 

\subsection{Summary}

This study compares two classical feature selection methods, forward and backward selection, against the taxonomy-based feature selection method. Three datasets, Arctic-Fox, AIS, and  Tropical Cyclones, are used for analysis. The approaches are tested using four models: logistic regression, random forest, XGBoost, and MLP. 
Seeding and stratified cross-validation are applied to maintain the reliability and validity of the results. Hyperparameter tuning is used to optimize the results further. 
The weighted-F1 score is used as the performance metric, and box plots are used to visualize the results.

\newpage

\section{Results and Analysis}
\label{ResultsAnalysis}
In this chapter, the results that are presented based on the experiments conducted according to the methodology discussed in Chapter 3 are presented. 
The box plots presented contain the baseline results for both forward and backward selection and the best feature subset result from our taxonomy method. 

\subsection{Arctic Fox dataset}
\label{FoxResult}

\begin{figure}[htbp]
    \centering
    \includegraphics[width=1.0\textwidth]{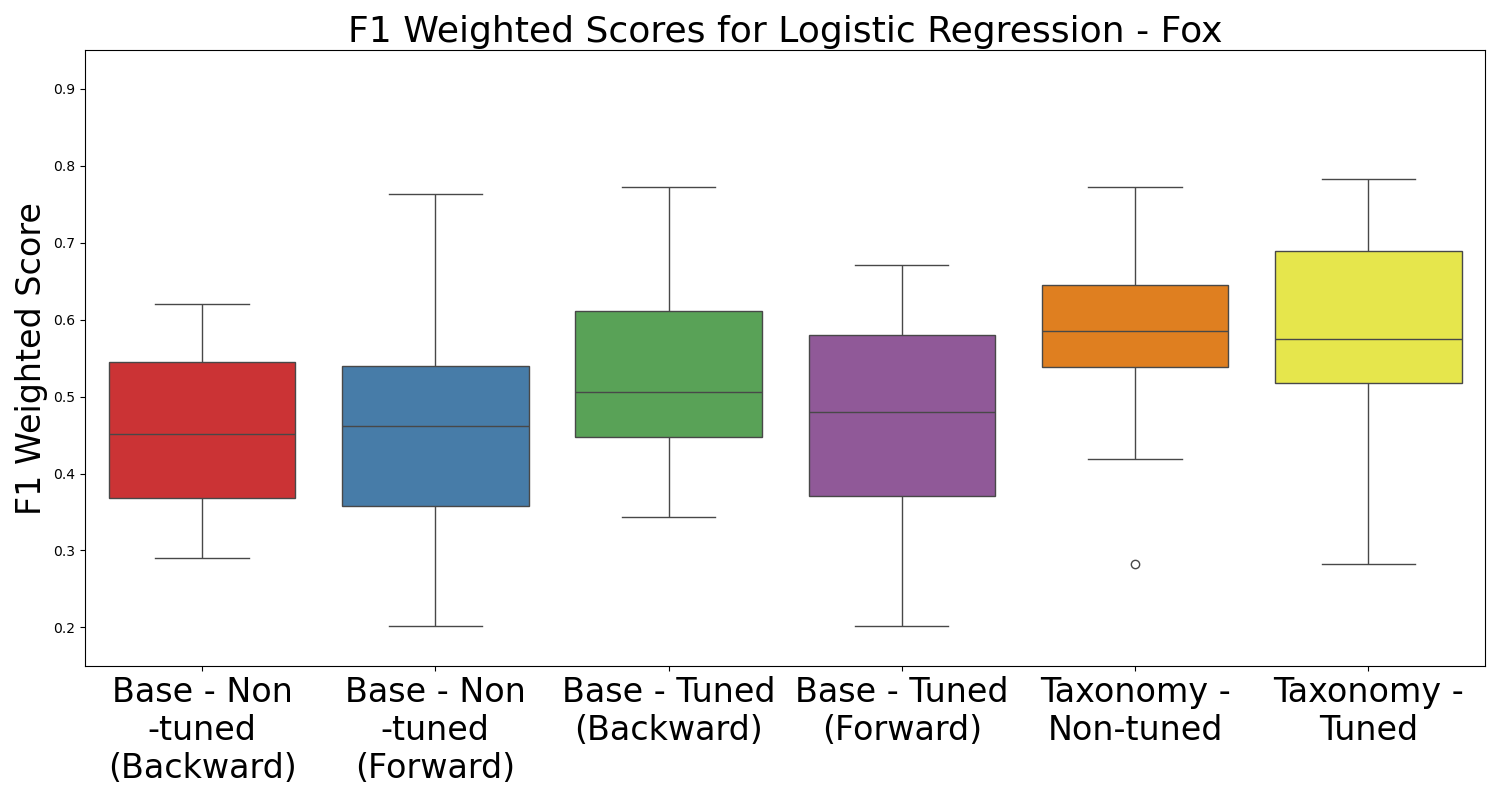}.
    \caption{Box-plot showing Weighted-F1 scores for different feature selection methods using Logistic Regression on the "Arctic Fox" dataset.}
    \label{fig:fox_logistic_1}
\end{figure}

\noindent Starting with logistic regression (Figure \ref{fig:fox_logistic_1}), the best baseline was a median Weighted-F1 score of $0.4615$, which was achieved by the forward selection method. 
It is important to highlight that this result is not better than random guessing, so this model performs very poorly.
The taxonomy method achieved a median Weighted-F1 score of $0.5852$, which is a significant improvement compared to other methods. 
This result was achieved through distance geometry (curvature), angles (indentation) from the taxonomy as its feature combination. 
Looking at the box plot, it is also evident that the taxonomy method has a tighter interquartile range, indicating that it was able to deliver more consistent results compared to the baseline methods.

In the tuned configuration, the best baseline result was achieved by backward selection with a median Weighted-F1 score of $0.5066$. 
While this is an improvement from the previous results, it's still far behind the non-tuned taxonomy result. 
However, the taxonomy with hyper-tuning reduced the median Weighted-F1 score to $0.5743$ and increased its interquartile range, reducing the method's consistent performance with the same feature combinations as the non-tuned set.

\begin{figure}[htbp]
    \centering
    \includegraphics[width=1.0\textwidth]{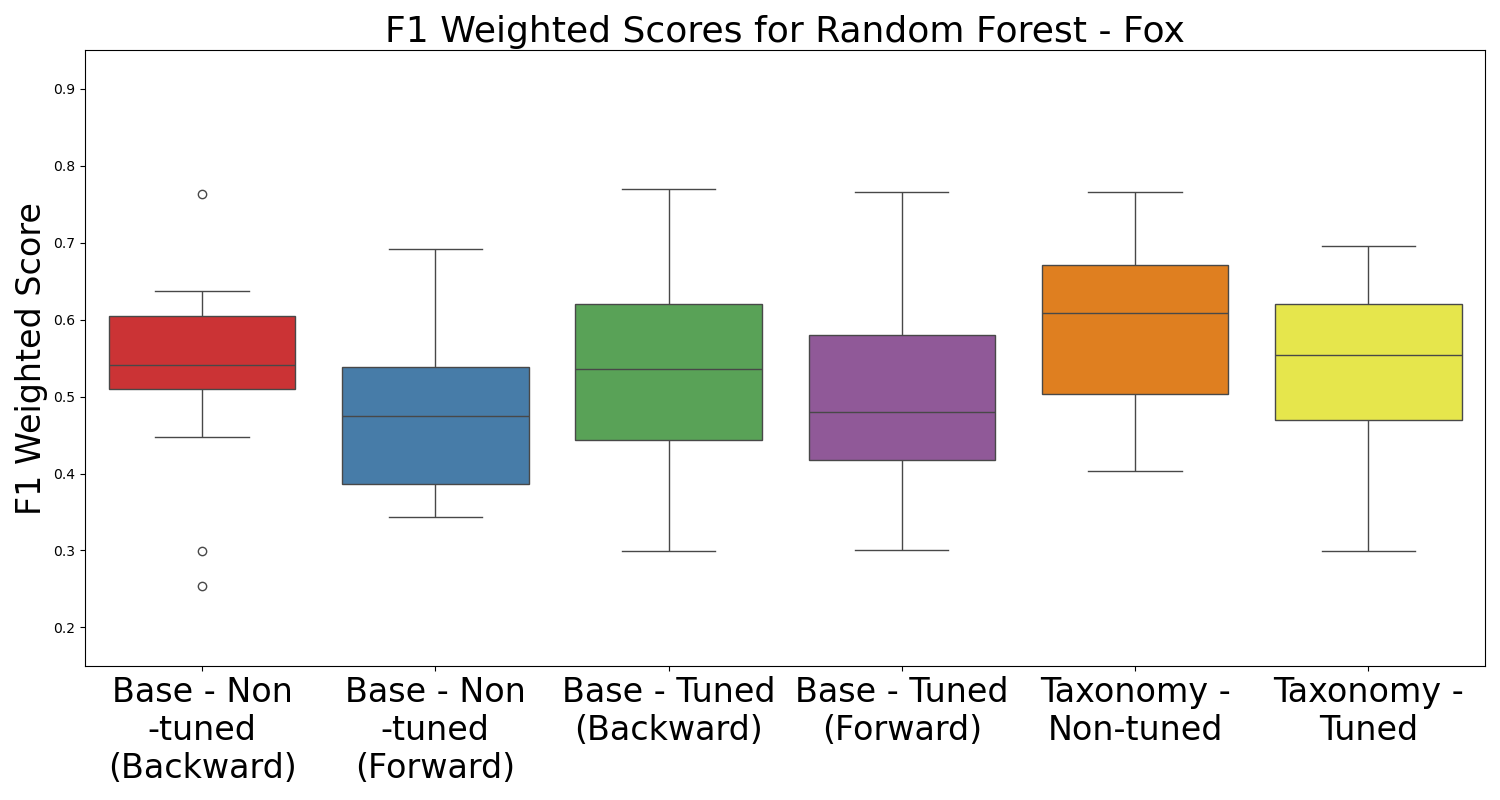}.
    \caption{Box-plot showing Weighted-F1 scores for different feature selection methods using Random Forest on the "Arctic Fox" dataset.}
    \label{fig:fox_rf_1}
\end{figure}

When evaluating random forests (Figure \ref{fig:fox_rf_1}), the best baseline non-tuned result was achieved by backward selection with a median Weighted-F1 score of $0.5412$. 
The taxonomy method produced a result of $0.6080$, which is higher than the baseline. 
After tuning, the best result was achieved by backward selection with a median Weighted-F1 score of $0.5357$, where the taxonomy method's results decreased to $0.5536$ even though it is still higher than the baseline. 
When looking at the consistency, the non-tuned backward selection showed the tightest interquartile ranges, however, including a few outliers. 
The interesting aspect of this result is that taxonomy only used acceleration-based features to produce a higher F1 score in both cases.

\begin{figure}[htbp]
    \centering
    \includegraphics[width=1.0\textwidth]{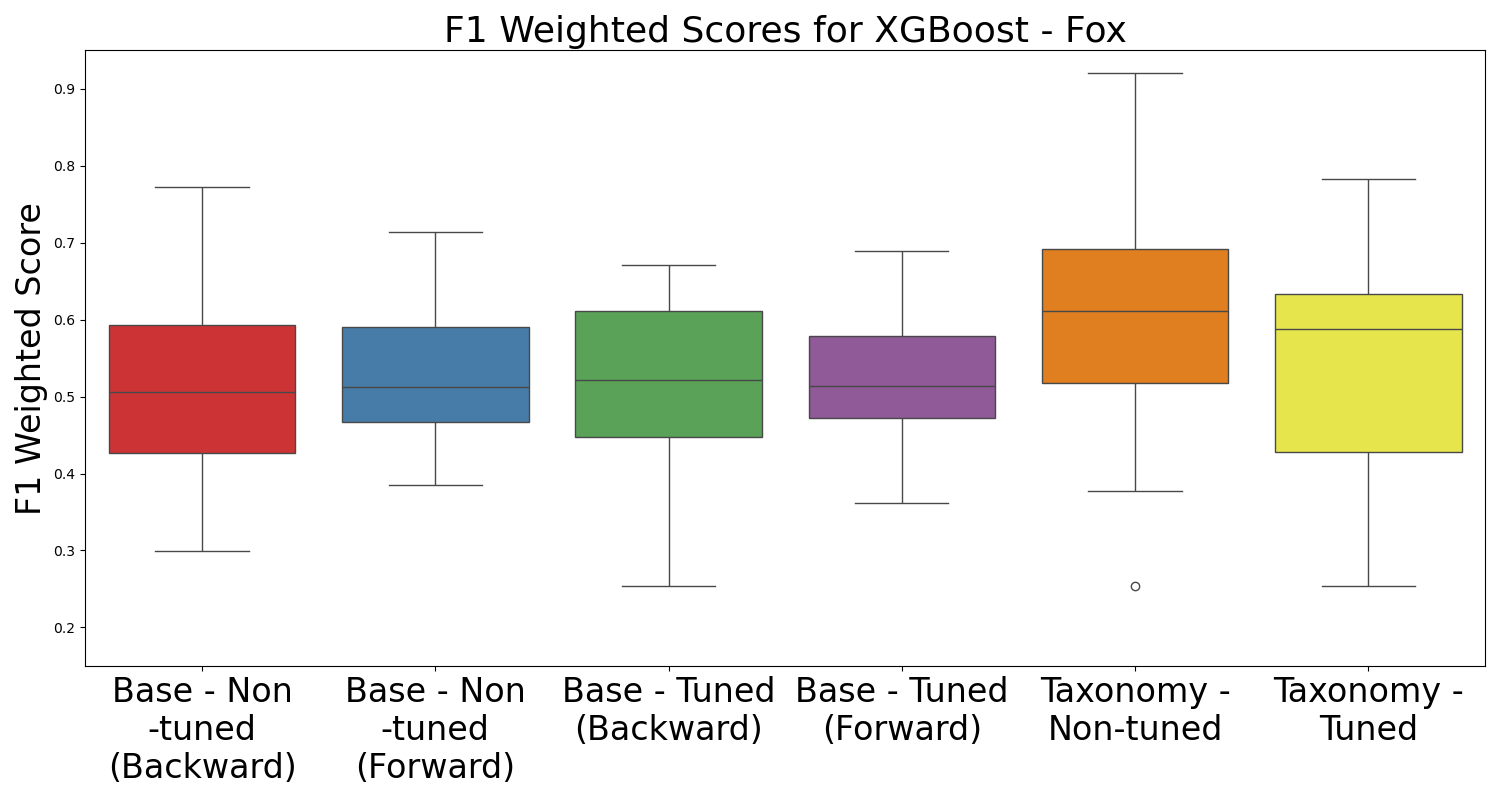}
    \caption{Box-plot showing Weighted-F1 scores for different feature selection methods using XGBoost on the "Arctic Fox" dataset.}
    \label{fig:fox_xgb_1}
\end{figure}

In the case of XGBoost (Figure \ref{fig:fox_xgb_1}), both baselines generated almost the same results in the non-tuned version; however, forward selection performed better with a median Weighted-F1 score of $0.5125$. 
The taxonomy method outperformed the baseline with a median Weighted-F1 score of $0.6111$ with distance geometry (curvature) and speed as its feature combination. 
When it comes to tuning, the baseline results did not change much except for the method itself, where backward selection had the highest median score of $0.5212$. 
The taxonomy results showed a decrease of $0.5884$, and acceleration was selected as the feature set. 
Looking at the overall plot, it is evident that the baseline showed much more consistent results with tighter inter-quartile ranges and no outliers than the taxonomy method in this specific context. 

\begin{figure}[htbp]
    \centering
    \includegraphics[width=1.0\textwidth]{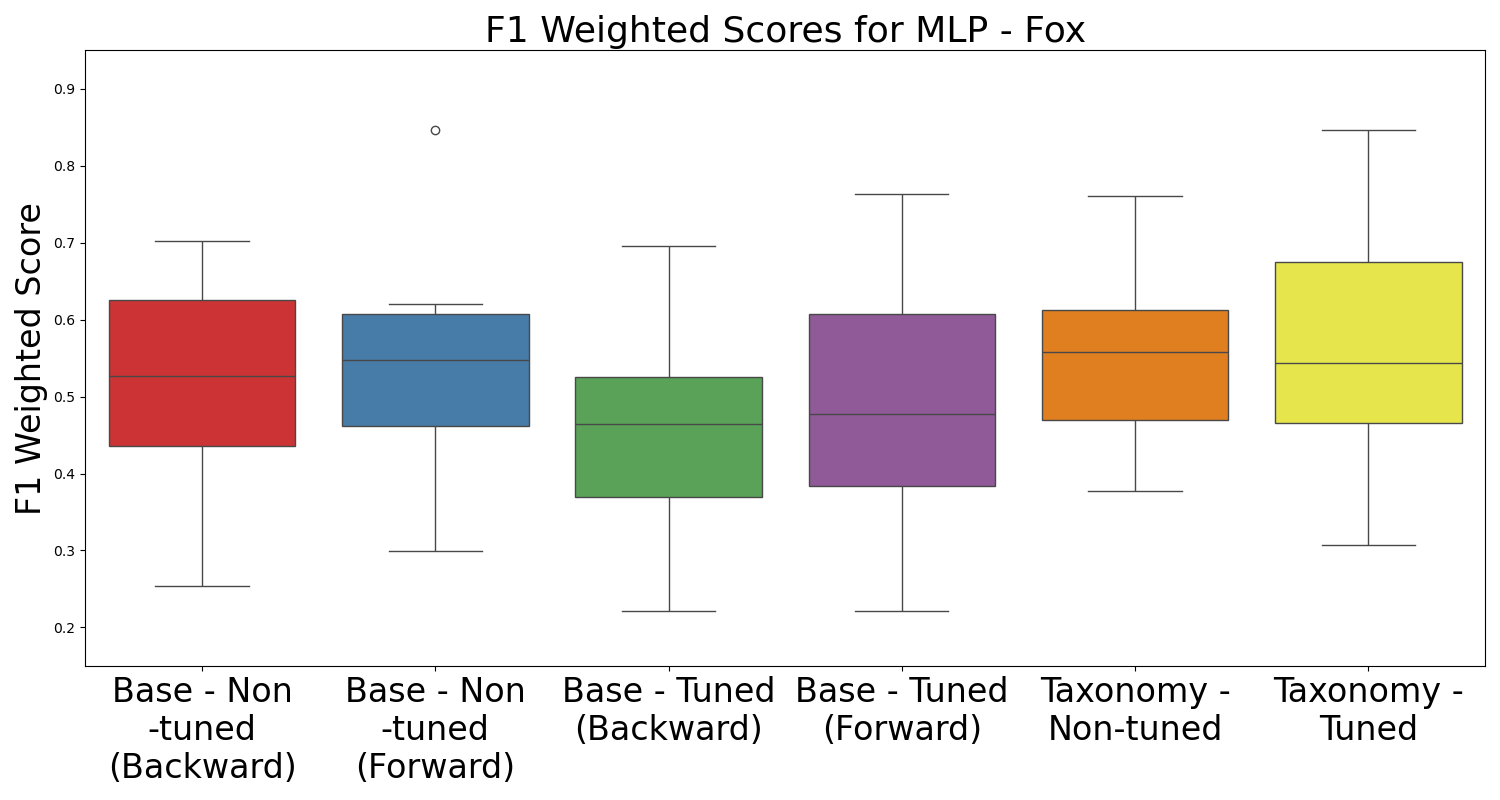}
    \caption{Box-plot showing Weighted-F1 scores for different feature selection methods using MLP on the "Arctic Fox" dataset.}
    \label{fig:fox_mlp_1}
\end{figure}

Finally, for the MLP model (Figure \ref{fig:fox_mlp_1}), the best non-tuned baseline result was achieved by the forward selection method, which produced a median Weighted-F1 score of $0.5482$. 
As for the taxonomy method, it produced a median Weighted-F1 score of $0.5440$, which slightly underperformed compared to the baselines with the distance geometry (curvature) and angles (indentation) as the feature combination. 
Looking at the tuned results, the forward selection method produced the best result of $0.4768$. 
In contrast, the taxonomy method produced a median Weighted-F1 score of $0.5579$, which is a bit higher than its non-tuned counterpart using the same feature combination. 
However, looking at the box plot, it is evident that it does increase the median and is able to visibly reduce the interquartile range,  improving the consistency of the results. 
Looking Table \ref{tab:fox_median_f1_1} that summarizes  the median Weighted-F1 score results for the Arctic Fox dataset, is evident that only the taxonomy-based method was able to gain a score above 0.6000. That too in two instances as such we can conclude in this  the case of the Arctic Fox dataset taxonomy-based feature selection was able to produce the better predictor.

\begin{table}[h!]
\centering
\caption{Median Weighted-F1 Scores for Arctic Fox Dataset}
\label{tab:fox_median_f1_1}
\begin{tabular}{lcccccc}
\toprule
\textbf{Model} & \multicolumn{2}{c}{\textbf{Forward}} & \multicolumn{2}{c}{\textbf{Backward}} & \multicolumn{2}{c}{\textbf{Taxonomy}} \\
\cmidrule(r){2-3} \cmidrule(r){4-5} \cmidrule(r){6-7}
 & \textbf{Non-Tuned} & \textbf{Tuned} & \textbf{Non-Tuned} & \textbf{Tuned} & \textbf{Non-Tuned} & \textbf{Tuned} \\
\midrule
Logistic Regression & 0.4615 & 0.4800   & 0.4512 & 0.5066 & 0.5852 & 0.5743 \\
Random Forest       & 0.4744 & 0.4800   & 0.5412 & 0.5357 & 0.6080  & 0.5536 \\
XG Boost            & 0.5125 & 0.5141 & 0.5064 & 0.5212 & 0.6111 & 0.5884 \\
MLP                 & 0.5482 & 0.4768 & 0.5271 & 0.4648 & 0.5440 & 0.5579 \\
\bottomrule
\end{tabular}
\end{table}

Next, Table \ref{tab:taxonomy_best_features_fox} summarizes the best performing taxonomy combination for each model in the context of the dataset.

\begin{table}[h!]
\centering
\caption{Best feature set for each model (taxonomy-based method)}
\label{tab:taxonomy_best_features_fox}
\begin{tabular}{ll}
\toprule
\textbf{Model} & \textbf{Best Feature Set} \\
\midrule
Logistic Regression - Non-Tuned & distance\_geometry (curvature) + angles (indentation) \\
Logistic Regression - Tuned     & distance\_geometry (curvature)  + angles (indentation)\\
MLP - Non-Tuned                 & distance\_geometry (curvature) + angles (indentation) \\
MLP - Tuned                     & angles (indentation) \\
Random Forest - Non-Tuned       & acceleration \\
Random Forest - Tuned           & acceleration \\
XGBoost - Non-Tuned             & distance\_geometry (curvature) + speed \\
XGBoost - Tuned                 & speed + acceleration \\
\bottomrule
\end{tabular}
\end{table}

Based on this table, a frequency analysis is carried out to explore the possibility of gaining added insight into the dataset. The results are visualized on Figure \ref{fig:fox_frequency}

\begin{figure}[htbp]
    \centering
    \includegraphics[width=1.0\textwidth]{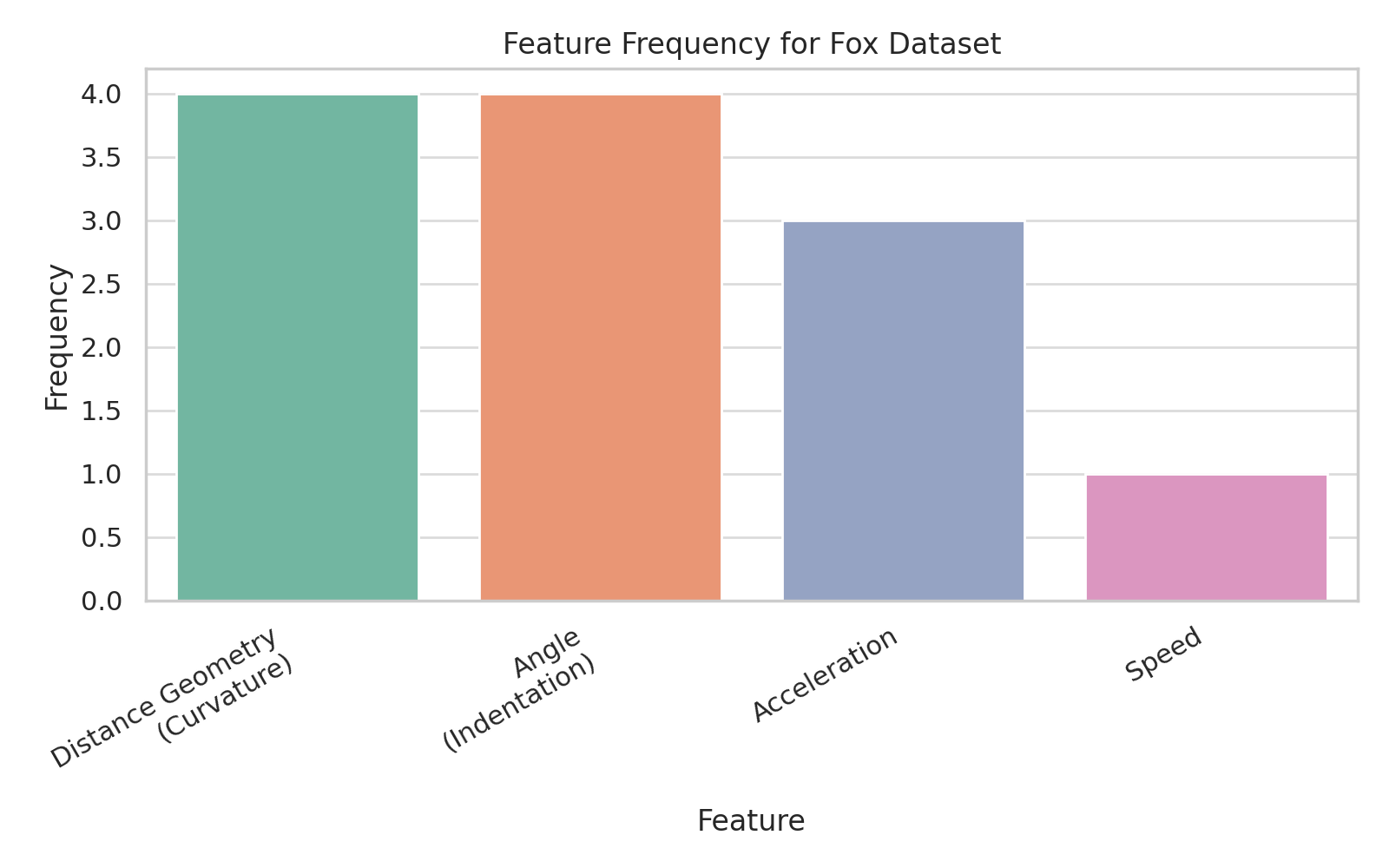}.
    \caption{Frequency of features}
    \label{fig:fox_frequency}
\end{figure}

Based on the above figure,it shows that distance\_geometry (curvature), angle (indentation) and acceleration are selected as features more frequently while speed has a much lower inclusion frequency. Thus, it shows that the data set is more sensitive to combinations that includes distance\_geometry (curvature), angle (indentation) and acceleration over speed.

\subsection{AIS dataset}

Figure \ref{fig:ais_logistic_1} shows the results for the AIS dataset with the logistic regression model.
Within the AIS dataset, the best baseline median Weighted-F1 score for logistic regression was $0.6968$, which was achieved by the forward selection method. 
The taxonomy method achieved a median Weighted-F1 of $0.6879$, marginally lower than other methods. 
The feature combination of angles (indentation), speed, and acceleration performance achieved this result. 
Looking at the box plot, the interquartile range for all the methods was similar, indicating similar consistency in the methods' performances.
In the tuned models, the best baseline result was still achieved by the forward selection method with a median Weighted-F1 score of $0.6928$; this is similar to the previous results with no improvement but still marginally above the non-tuned taxonomy result. However, the taxonomy with hyper-tuning reduced the median Weighted-F1 score to $0.6779$, but the interquartile range became smaller, thereby improving the consistency. The selected feature combination changed from angles (indentation), speed, and acceleration to distance geometry (curvature), angles (indentation), and speed combination.

\begin{figure}[htbp]
    \centering
    \includegraphics[width=1.0\textwidth]{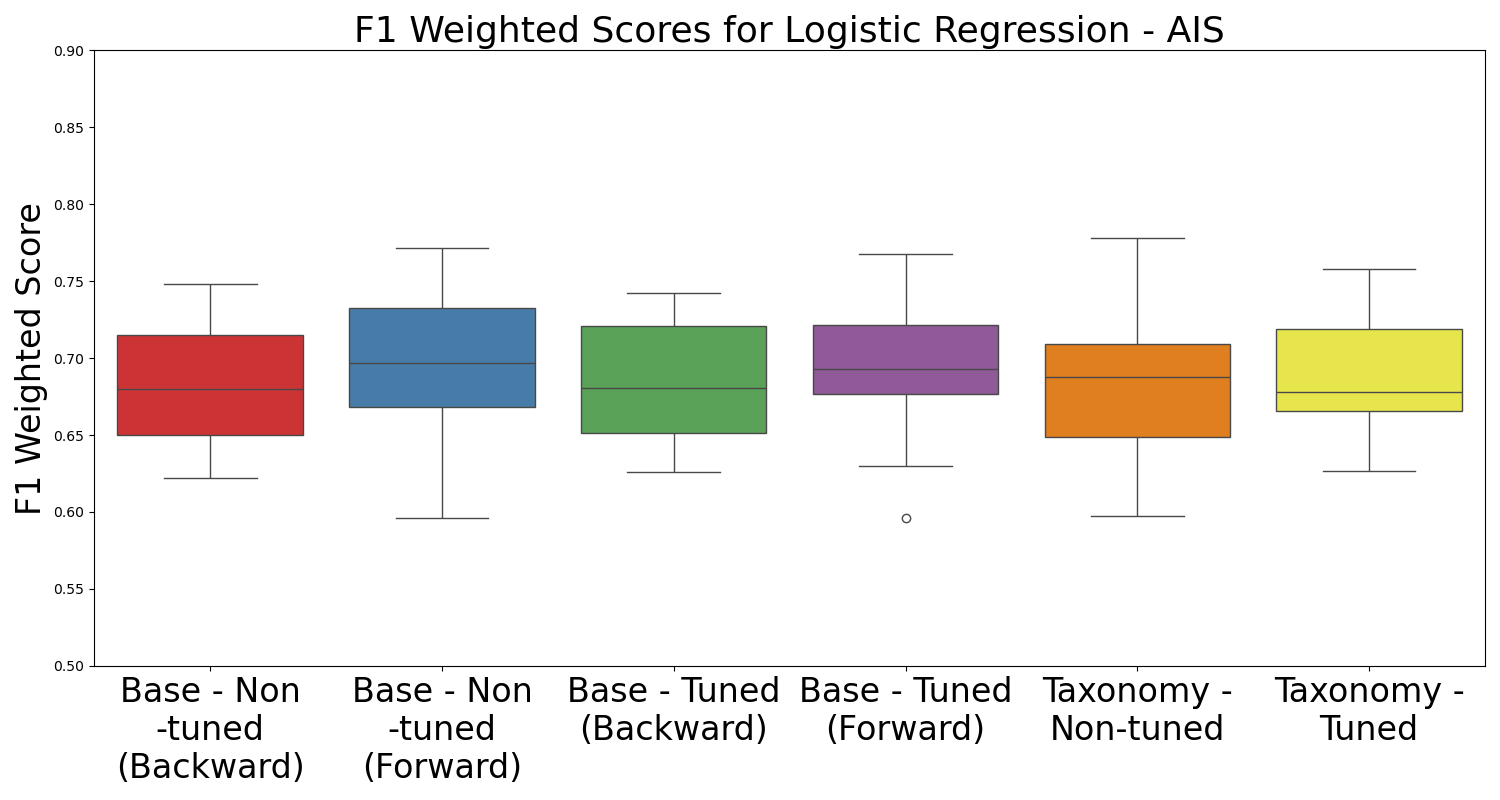}.
    \caption{Box-plot showing Weighted-F1 scores for different feature selection methods using Logistic Regression on the "AIS" dataset.}
    \label{fig:ais_logistic_1}
\end{figure}

While evaluating the random forest (Figure \ref{fig:ais_rf}), the best baseline non-tuned result was achieved by forward selection with a median Weighted-F1 score of $0.7516$. In contrast, the taxonomy method produced a median Weighted-F1 score of $0.7598$ from the combination of distance geometry (curvature), speed, and acceleration features. That was a slight improvement in comparison to the non-tuned baseline results. Whereas, after hyper-tuning, the best median Weighted-F1 score was still the same $0.7516$, as non-tuned, which was still achieved by the forward selection method. The taxonomy method’s result decreased to $0.7428$, a slight decrease from the baseline scores. Along with the scores, there was variation in the combination of features that was instead of acceleration. Now, angles (indentation) were selected in combination with distance geometry (curvature) and speed. While looking at the consistency, the non-tuned methods showed the tightest interquartile ranges, and there were no outliers.

\begin{figure}[htbp]
    \centering
    \includegraphics[width=1.0\textwidth]{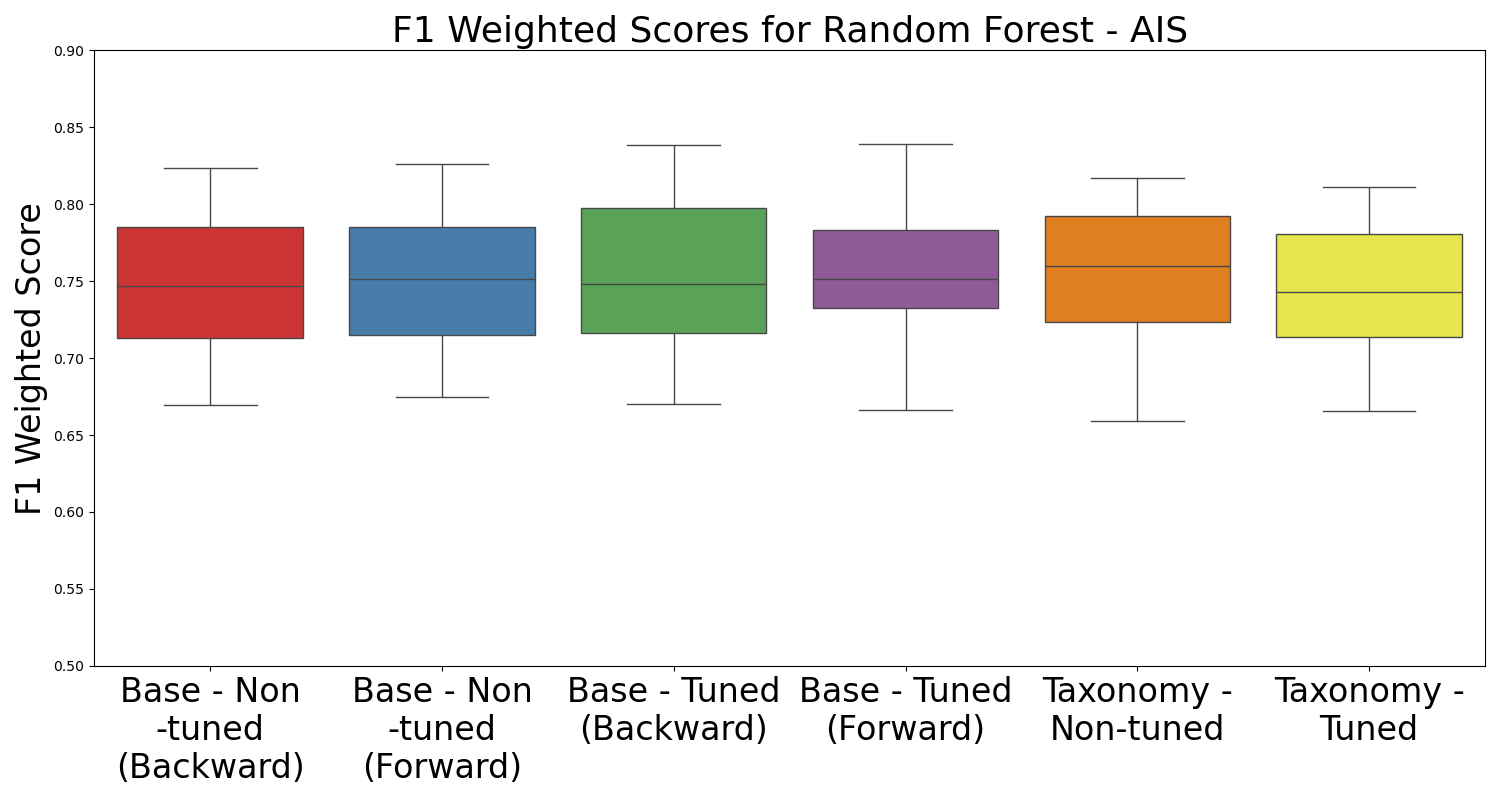}
    \caption{Box-plot showing Weighted-F1 scores for different feature selection methods using Random Forest on the "AIS" dataset.}
    \label{fig:ais_rf}
\end{figure}

In evaluating the XGBoost model (Figure \ref{fig:ais_xgb}), the best baseline non-tuned result was achieved by backward selection with a median Weighted-F1 score of $0.7381$. 
Here, the taxonomy method achieved a slightly improved score of $0.7438$ with the feature combinations of distance geometry (curvature), angles (indentation), speed, and acceleration. 
When the baselines were tuned, their results improved. 
The best method after tuning was backward selection with the median Weighted-F1 score of 0.7597. 
There was an increase in tuned results of the taxonomy method with the median Weighted-F1 score of $0.7535$. 
However, it is marginally lower than the tuned baseline results, with the feature combination remaining the same four features. 
When looking at the box-plots, the overall interquartile ranges are tighter and consistent, and all are above a score of $0.7000$.

\begin{figure}[htbp]
    \centering
    \includegraphics[width=1.0\textwidth]{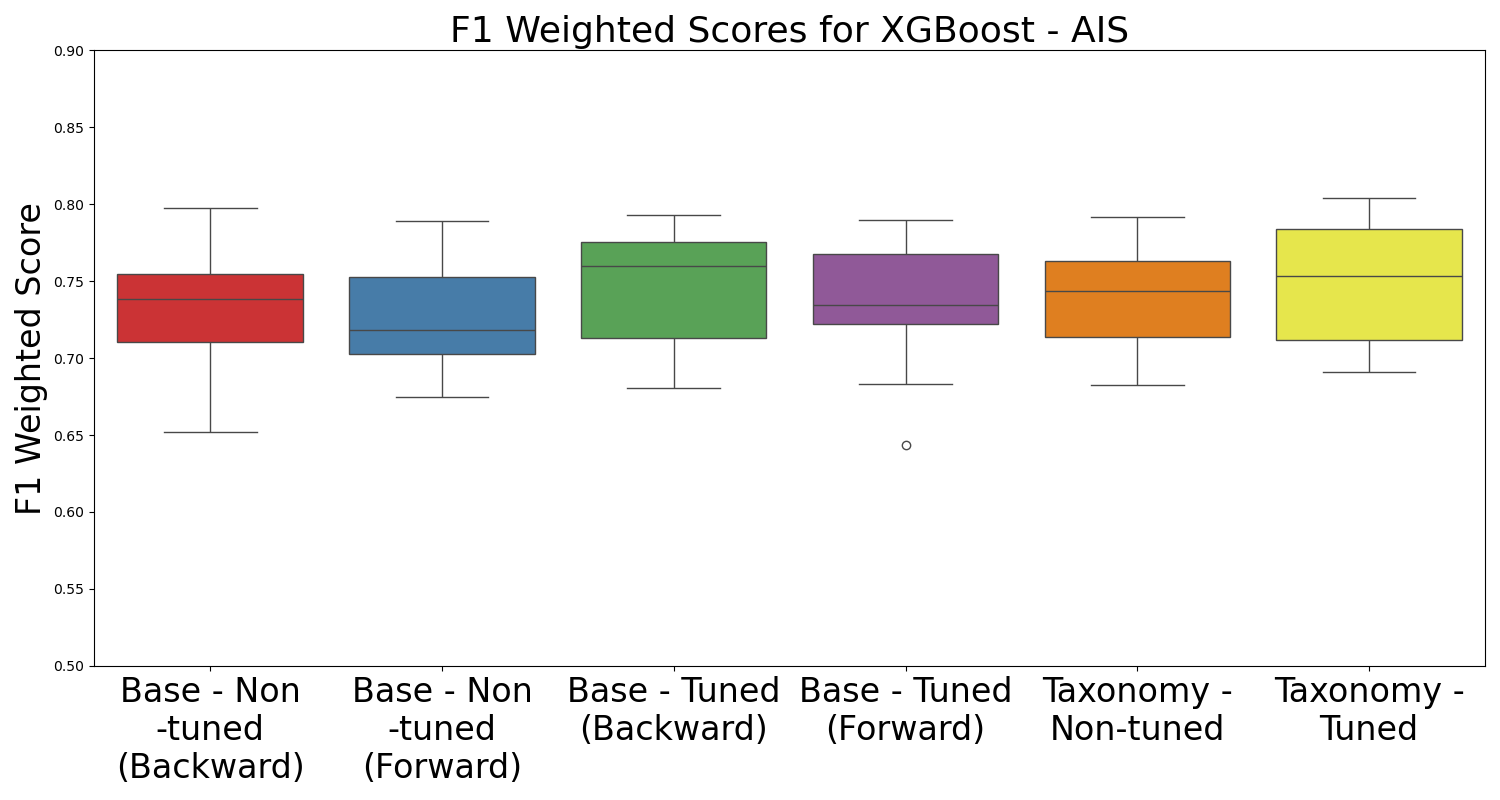}
    \caption{Box-plot showing Weighted-F1 scores for different feature selection methods using XGBoost on the "AIS" dataset.}
    \label{fig:ais_xgb}
\end{figure}

Finally, while evaluating the MLP model (Figure \ref{fig:ais_mlp}), the best non-tuned baseline was achieved by the forward selection method with a median Weighted-F1 score of $0.6367$. 
In comparison, the taxonomy method achieved far better results, with an increase in the median Weighted-F1 score being $0.6711$, through the feature combination of speed and acceleration. 
After hyperparameter tuning, the forward selection performed better, with a noticeable increase in the median Weighted-F1 score being $0.7028$. 
The taxonomy median Weighted-F1 score was $0.7095$, the highest result among all other methods. To achieve these results, the feature combination included angle (indentation) along with speed and acceleration when compared to a non-tuned taxonomy feature combination. 
By looking at the box plots, the interquartile range for non-tuned taxonomy was tighter with few outliers. Overall, the taxonomy-based method performed better than its counterparts.

\begin{figure}[htbp]
    \centering
    \includegraphics[width=1.0\textwidth]{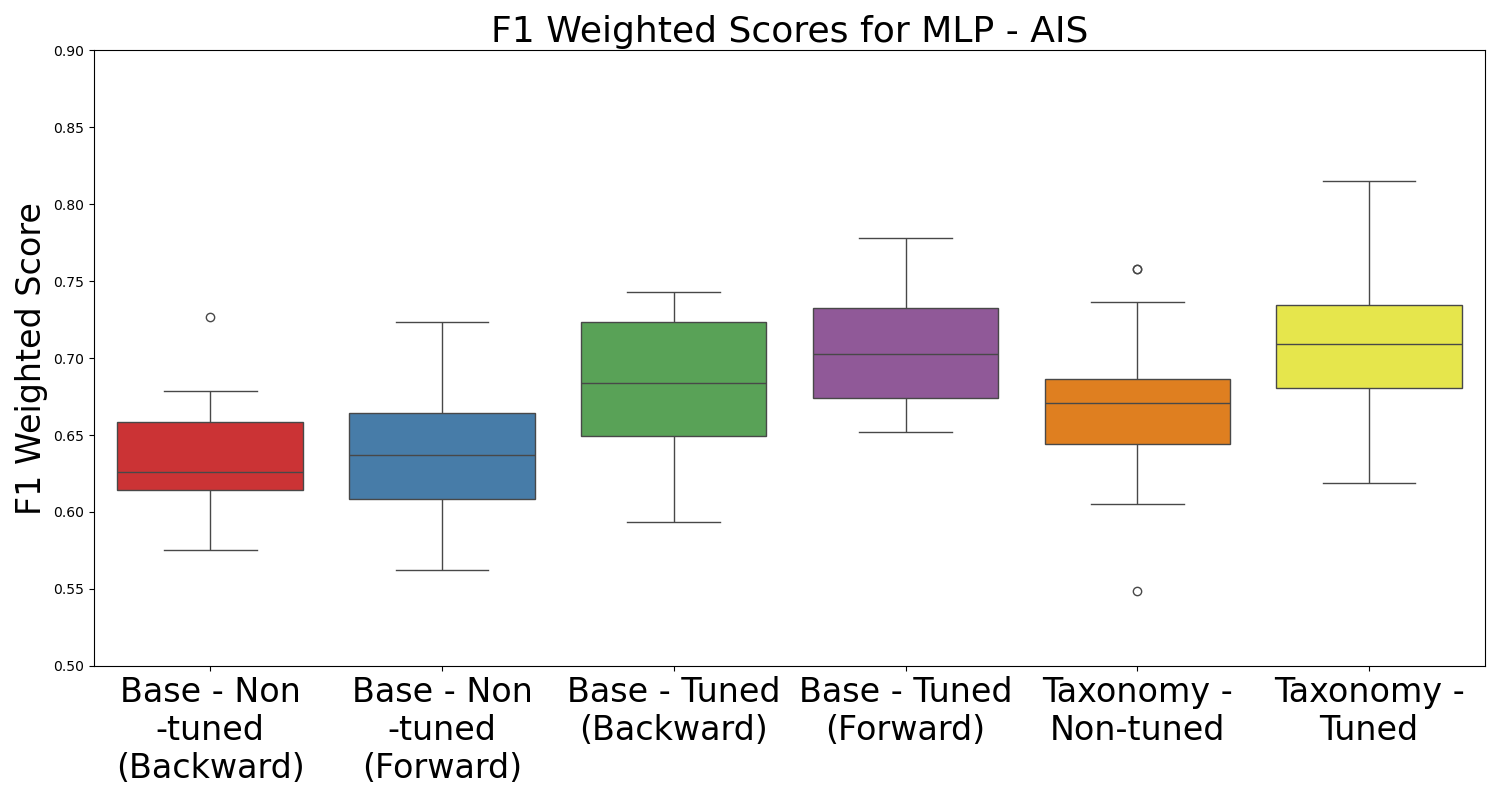}
    \caption{Box-plot showing Weighted-F1 scores for different feature selection methods using MLP on the "AIS" dataset.}
    \label{fig:ais_mlp}
\end{figure}

Overall, the performances in this dataset was comparatively way better as all the results were above $0.6000$. 
Looking at the Table \ref{tab:AIS_median_f1} that summarizes the median weighted-F1 score results for the AIS Dataset, it is evident that taxonomy-based feature selection was able to produce better or at least comparable results. One interesting result is that in the non-tuned scores in all cases except one taxonomy-based method produced produced better results. However the impact of hyperparameters seem to be minimal or even negative for except the MLP model where there is a considerable improvement. 

\begin{table}[h!]
\centering
\caption{Median Weighted-F1 Scores for AIS Dataset}
\label{tab:AIS_median_f1}
\begin{tabular}{lcccccc}
\toprule
\textbf{Model} & \multicolumn{2}{c}{\textbf{Forward}} & \multicolumn{2}{c}{\textbf{Backward}} & \multicolumn{2}{c}{\textbf{Taxonomy}} \\
\cmidrule(r){2-3} \cmidrule(r){4-5} \cmidrule(r){6-7}
 & \textbf{Non-Tuned} & \textbf{Tuned} & \textbf{Non-Tuned} & \textbf{Tuned} & \textbf{Non-Tuned} & \textbf{Tuned} \\
\midrule
Logistic Regression & 0.6968 & 0.6928 & 0.6801 & 0.6806 & 0.6879 & 0.6779 \\
Random Forest       & 0.7516 & 0.7516 & 0.7471 & 0.7483 & 0.7598 & 0.7428 \\
XG Boost            & 0.7183 & 0.7345 & 0.7381 & 0.7597 & 0.7438 & 0.7535 \\
MLP                 & 0.6367 & 0.7028 & 0.6261 & 0.6837 & 0.6711 & 0.7095 \\
\bottomrule
\end{tabular}
\end{table}

Next, Table \ref{tab:taxonomy_best_features_ais} summaries the best performing taxonomy combination for each model in the context of the dataset. Based on this table, a frequency analysis is carried out to explore the possibility of gaining added insight into the dataset. The results are visualized on Figure \ref{fig:ais_frequency}

\begin{table}[h!]
\centering
\caption{Best feature set for each model on AIS dataset (taxonomy-based method)}
\label{tab:taxonomy_best_features_ais}
\begin{tabular}{l p{9cm}}
\toprule
\textbf{Model} & \textbf{Best Feature Set} \\
\midrule
Logistic Regression - Non-Tuned & angles (indentation) + speed + acceleration \\
Logistic Regression - Tuned     & distance\_geometry (curvature) + angles (indentation) + speed \\
MLP - Non-Tuned                 & speed + acceleration \\
MLP - Tuned                     & angles (indentation) + speed + acceleration \\
Random Forest - Non-Tuned       & distance\_geometry (curvature) + speed + acceleration \\
Random Forest - Tuned           & distance\_geometry (curvature) + speed + acceleration \\
XGBoost - Non-Tuned             & distance\_geometry (curvature) + speed + acceleration \\
XGBoost - Tuned                 & distance\_geometry (curvature) + angles (indentation) + speed + acceleration \\
\bottomrule
\end{tabular}
\end{table}

\begin{figure}[htbp]
    \centering
    \includegraphics[width=1.0\textwidth]{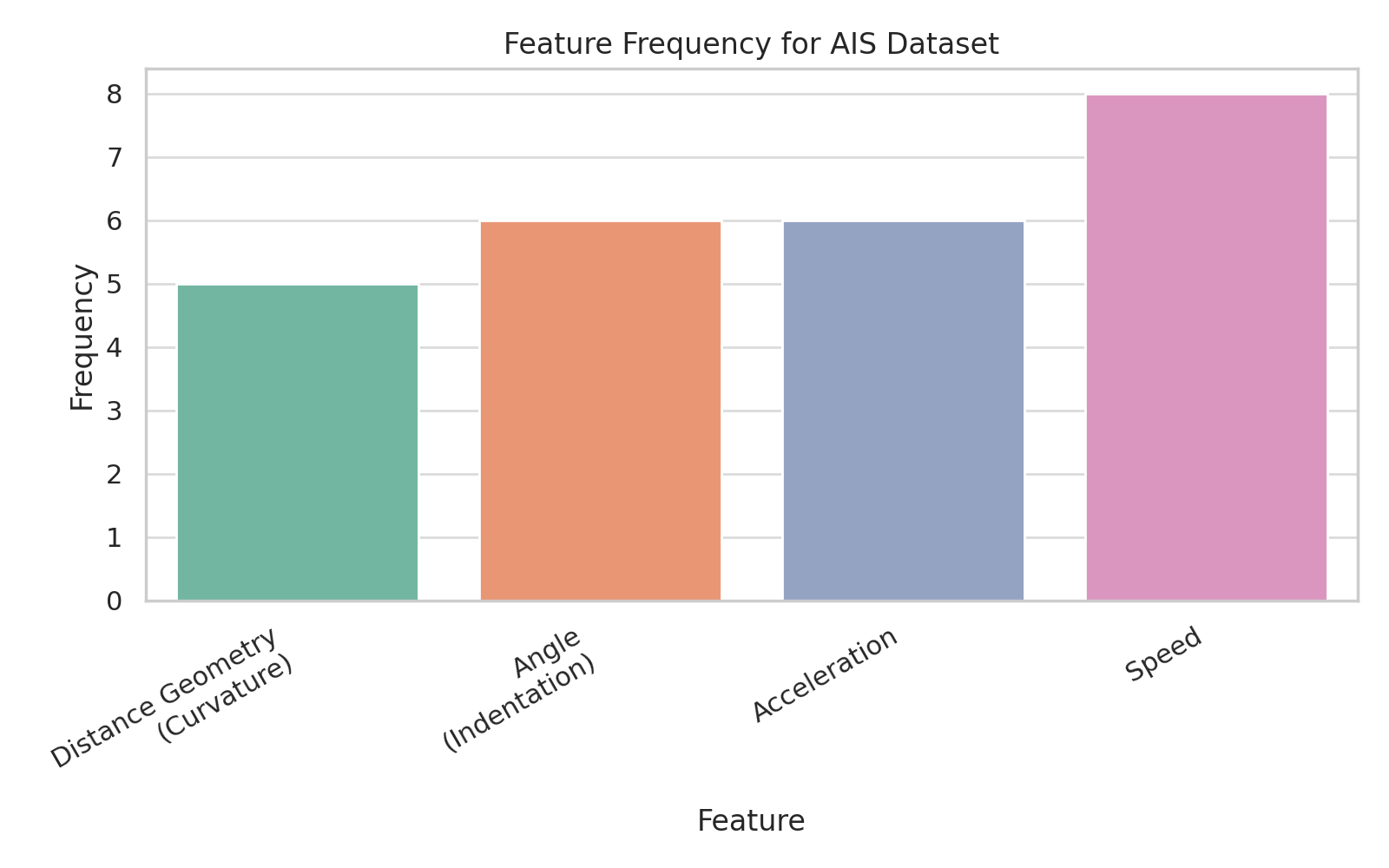}.
    \caption{Frequency of features}
    \label{fig:ais_frequency}
\end{figure}

The results suggests that all feature categories were used for the best combination similarly however, with a slight inclination towards speed. This indicates that the most of the feature categories are similarly important when classifying the AIS dataset in turn should be considered in making decisions related to this dataset.

\subsection{Tropical Cyclone dataset}

\begin{figure}[htbp]
    \centering
    \includegraphics[width=1.0\textwidth]{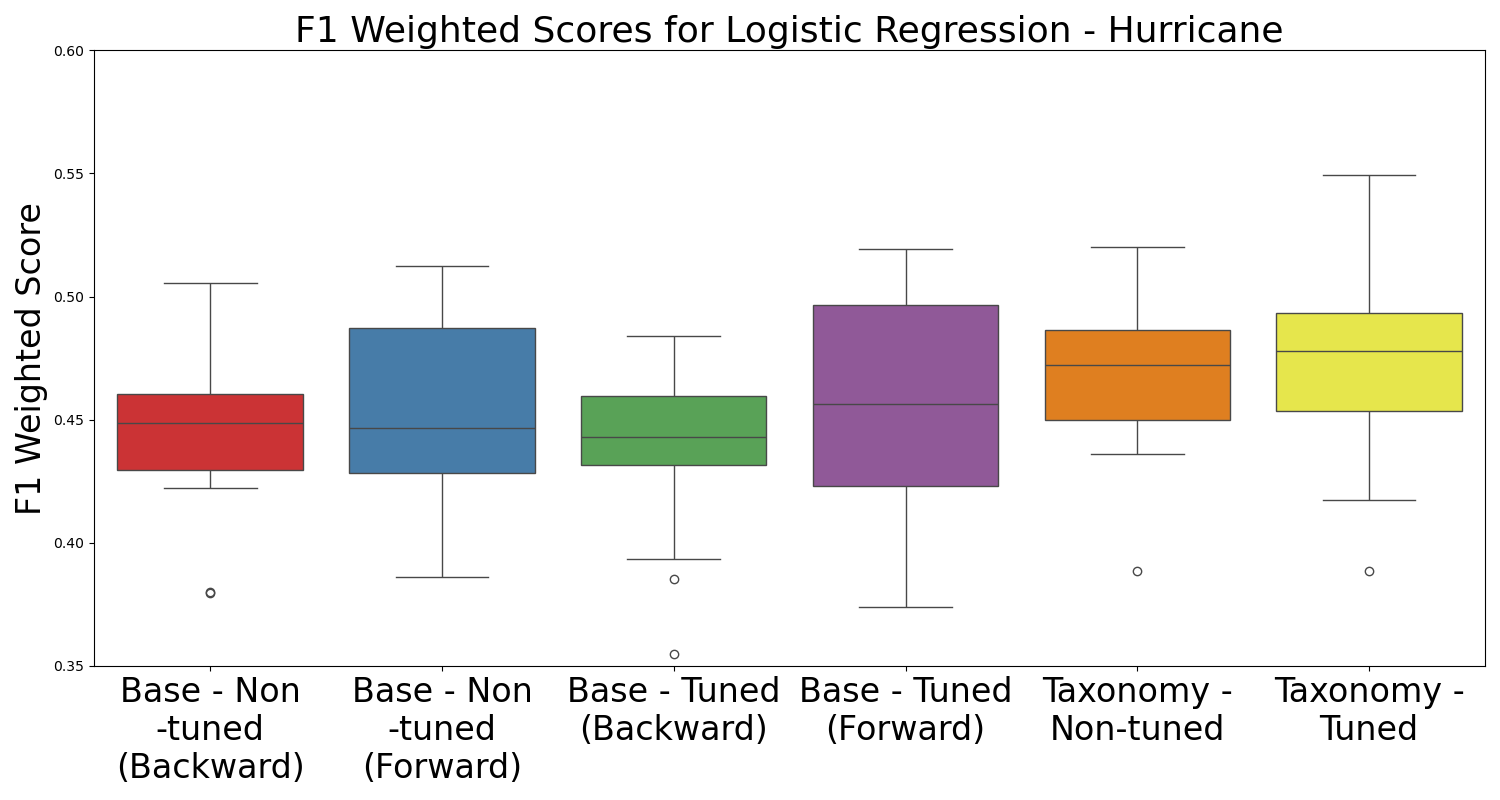}.
    \caption{Box-plot showing Weighted-F1 scores for different feature selection methods using Logistic Regression on the "Hurricane" dataset.}
    \label{fig:hurr_logistic}
\end{figure}

In the context of the Tropical Cyclone dataset and the logistic regression model (Figure \ref{fig:hurr_logistic}), backward feature selection scores the best result with a median Weighted-F1 score of $0.4487$. 
At the same time, the taxonomy method outperformed it with a median Weighted-F1 score of $0.4723$ with the feature combination distance geometry (curvature), angle (indentation), speed, and acceleration. In the tuned results, the forward selection method scored the best results for the baseline with a median Weighted-F1 score of $0.4565$. 
In contrast, the taxonomy method outperformed the baseline results with a median Weighted-F1 score of $0.4779$ with the same feature combination as the non-tuned feature set. 
Even though the taxonomy model outperformed the baseline, the overall results of the models were lacking overall. 
Looking at the interquartile range, the backward and the taxonomy approach seem to deliver much more consistent results compared to forward selection.

\begin{figure}[htbp]
    \centering
    \includegraphics[width=1.0\textwidth]{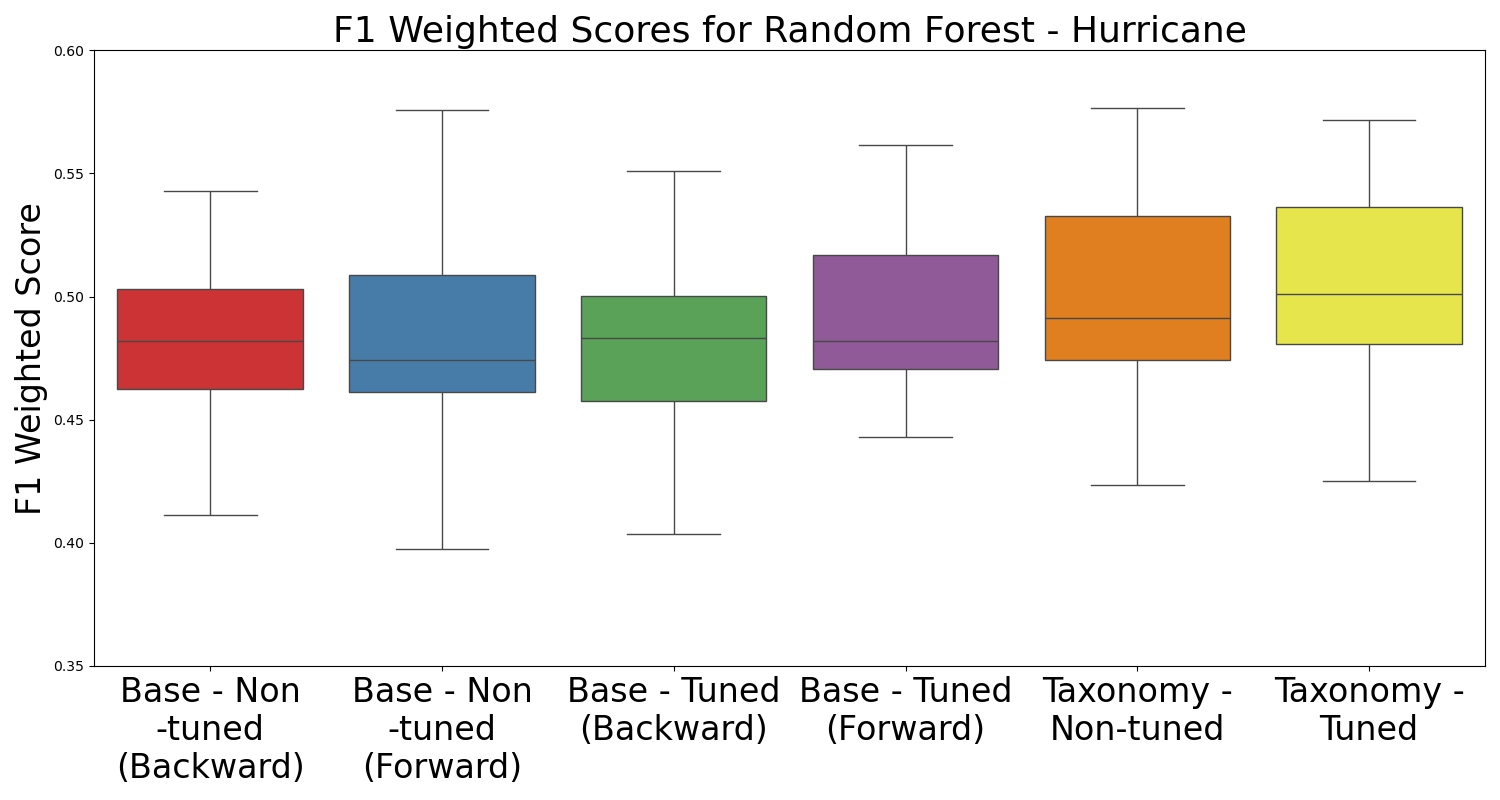}
    \caption{Box-plot showing Weighted-F1 scores for different feature selection methods using Random Forest on the "Hurricane" dataset.}
    \label{fig:hurr_rf}
\end{figure}

When random forest is considered (Figure \ref{fig:hurr_rf}), the backward selection method with a median Weighted-F1 score of $0.4820$ was the best non-tuned result that was registered for the baseline. 
Next, the non-tuned taxonomy-based method registered a median Weighted-F1 score of $ 0.4913$, which shows a slight improvement with the feature combination of distance geometry (curvature), angles (indentation), speed, and acceleration. 
After hyperparameter tuning, the backward selection method performed the best with a median Weighted-F1 score of $0.4832$. 
The taxonomy approach registered a median Weighted-F1 score of $0.5009$ with the same combination of features from the non-tuned feature set. Looking at the box plots, the overall interquartile ranges are similar, with all methods indicating similar consistency among them.

\begin{figure}[htbp]
    \centering
    \includegraphics[width=1.0\textwidth]{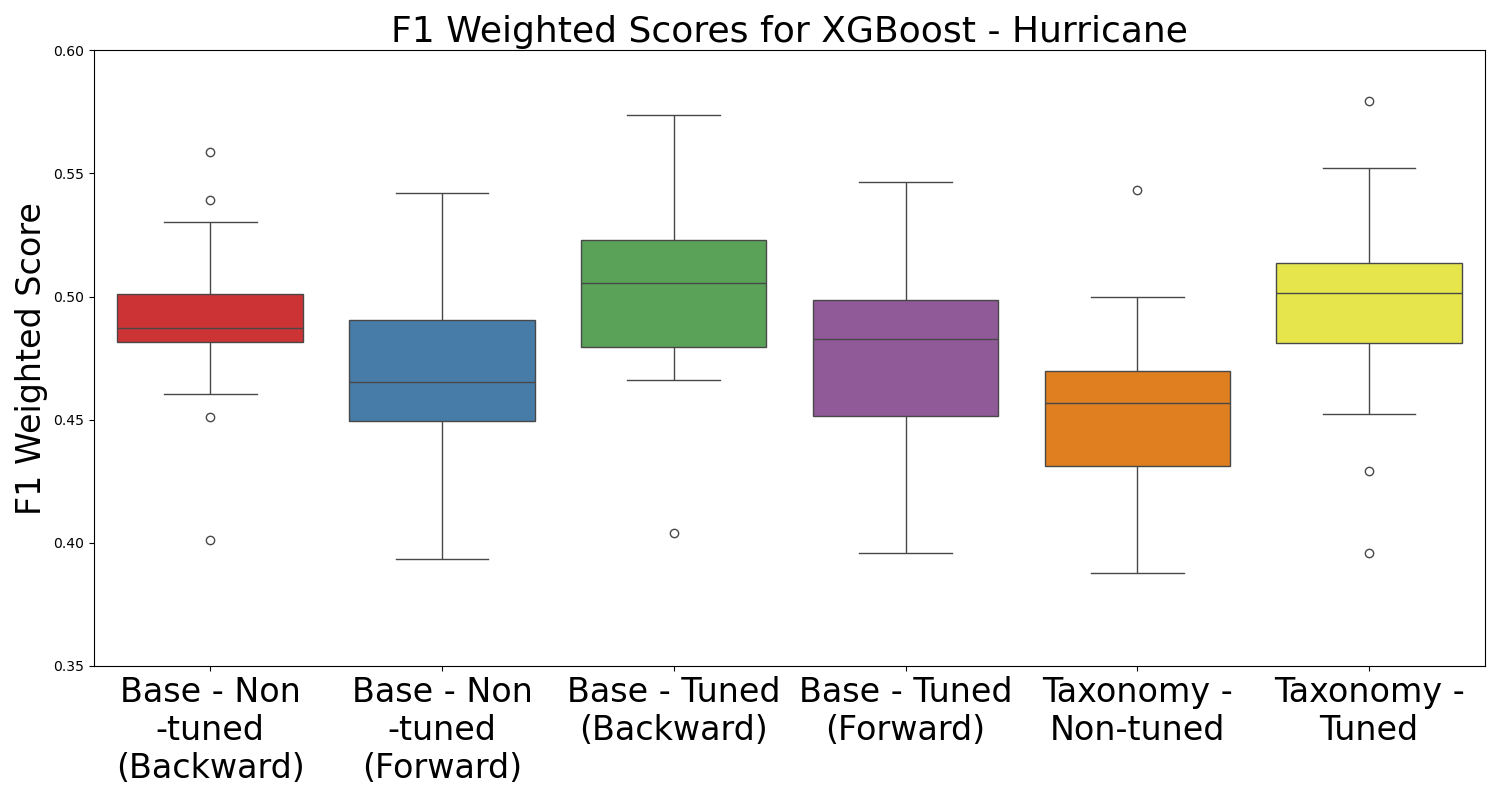}
    \caption{Box-plot showing Weighted-F1 scores for different feature selection methods using XGBoost on the "Hurricane" dataset.}
    \label{fig:hurr_xgb}
\end{figure}

When evaluating the XGBoost model (Figure \ref{fig:hurr_xgb}), backward selection registered the best results with a median Weighted-F1 score of $0.4873$. 
However, taxonomy underperformed compared to the baseline with a median Weighted-F1 score of $0.4567$ with distance geometry (curvature), angle (indentation), and speed as its feature combination. 
After hyperparameter tuning, the backward selection method performed the best for the baseline with a $0.5055$, whereas the taxonomy-based method registered a similar result of $0.5014$ with the feature combination distance geometry (curvature), angle (indentation), acceleration, and speed. 
The interquartile ranges seem similar for all except the non-tuned backward selection method, where it showed a much tighter range, indicating it had the most consistent results.

\begin{figure}[htbp]
    \centering
    \includegraphics[width=1.0\textwidth]{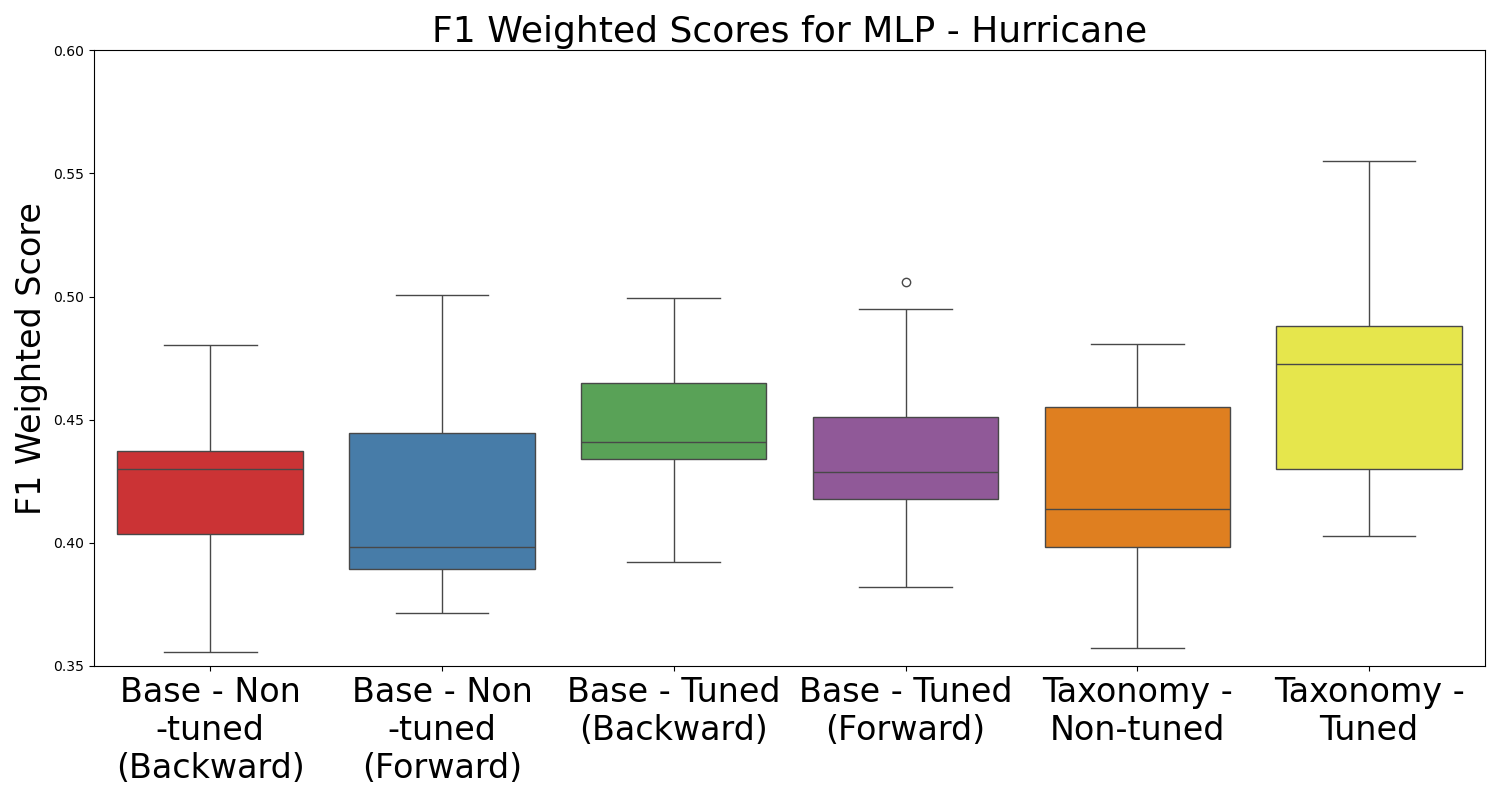}
    \caption{Box-plot showing Weighted-F1 scores for different feature selection methods using MLP on the "Hurricane" dataset.}
    \label{fig:fox_mlp}
\end{figure}

Finally, while evaluating the MLP model, the backward selection method registered the best result of a median Weighted-F1 score of $0.4301$. 
The taxonomy method registered a median Weighted-F1 score of $0.4136$, which is lower than the baseline with the feature combination set angle (indentation), speed, and acceleration. 
After performing hyperparameter tuning, the backward selection method performed the best for the baseline at $0.4411$. 
Whereas the taxonomy-based method performed better with a median Weighted-F1 score of $0.4725$ with the feature combination of speed and acceleration to register this result. 
By looking at the box plot, the tuned taxonomy method performed better, but when interquartile ranges were compared, they were tighter and more consistent for tuned and non-tuned backward selection methods. 

This dataset seems to be more difficult to classify overall, with all results being under 0.5100. This could be because the samples were selected from the pool of data that ranged over 200 years, resulting in a dataset that is difficult to classify with accuracy. 

Looking at the Table \ref{tab:cyclone_median_f1} that summarizes the median Weighted-F1 score results for the Tropical Cyclone Dataset, it is evident that none of the feature selection methods preform to at satisfactory  level . However in the non-tuned results, the taxonomy-based method seems to outperform the classical methods in all cases but one. 

\begin{table}[h!]
\centering
\caption{Median Weighted-F1 Scores for Tropical Cyclone Dataset}
\label{tab:cyclone_median_f1}
\begin{tabular}{lcccccc}
\toprule
\textbf{Model} & \multicolumn{2}{c}{\textbf{Forward}} & \multicolumn{2}{c}{\textbf{Backward}} & \multicolumn{2}{c}{\textbf{Taxonomy}} \\
\cmidrule(r){2-3} \cmidrule(r){4-5} \cmidrule(r){6-7}
 & \textbf{Non-Tuned} & \textbf{Tuned} & \textbf{Non-Tuned} & \textbf{Tuned} & \textbf{Non-Tuned} & \textbf{Tuned} \\
\midrule
Logistic Regression & 0.4464 & 0.4565 & 0.4487 & 0.4429 & 0.4723 & 0.4779 \\
Random Forest       & 0.4742 & 0.4820 & 0.4820 & 0.4832 & 0.4913 & 0.5009 \\
XG Boost            & 0.4655 & 0.4827 & 0.4873 & 0.5055 & 0.4567 & 0.5014 \\
MLP                 & 0.3984 & 0.4286 & 0.4301 & 0.4411 & 0.4136 & 0.4725 \\
\bottomrule
\end{tabular}
\end{table}

Next, Table \ref{tab:taxonomy_best_features_tc} summarizes the best-performing taxonomy combination for each model in the context of the dataset. Based on this table, a frequency analysis is carried out to explore the possibility of gaining added insight into the dataset. The results are visualized on Figure \ref{fig:tc_frequency}

\begin{table}[h!]
\centering
\caption{Best feature set for each model on Tropical Cyclone dataset (taxonomy-based method)}
\label{tab:taxonomy_best_features_tc}
\begin{tabular}{l p{9cm}}
\toprule
\textbf{Model} & \textbf{Best Feature Set} \\
\midrule
Logistic Regression - Non-Tuned & distance\_geometry (curvature) + angles (indentation) + speed + acceleration \\
Logistic Regression - Tuned     & distance\_geometry (curvature) + angles (indentation) + speed + acceleration \\
MLP - Non-Tuned                 & angles (indentation) + speed + acceleration \\
MLP - Tuned                     & speed + acceleration \\
Random Forest - Non-Tuned       & distance\_geometry (curvature) + angles (indentation) + speed + acceleration \\
Random Forest - Tuned           & distance\_geometry (curvature) + angles (indentation) + speed + acceleration \\
XGBoost - Non-Tuned             & distance\_geometry (curvature) + angles (indentation) + speed + acceleration \\
XGBoost - Tuned                 & distance\_geometry (curvature) + angles (indentation) + speed + acceleration \\
\bottomrule
\end{tabular}
\end{table}

\begin{figure}[htbp]
    \centering
    \includegraphics[width=1.0\textwidth]{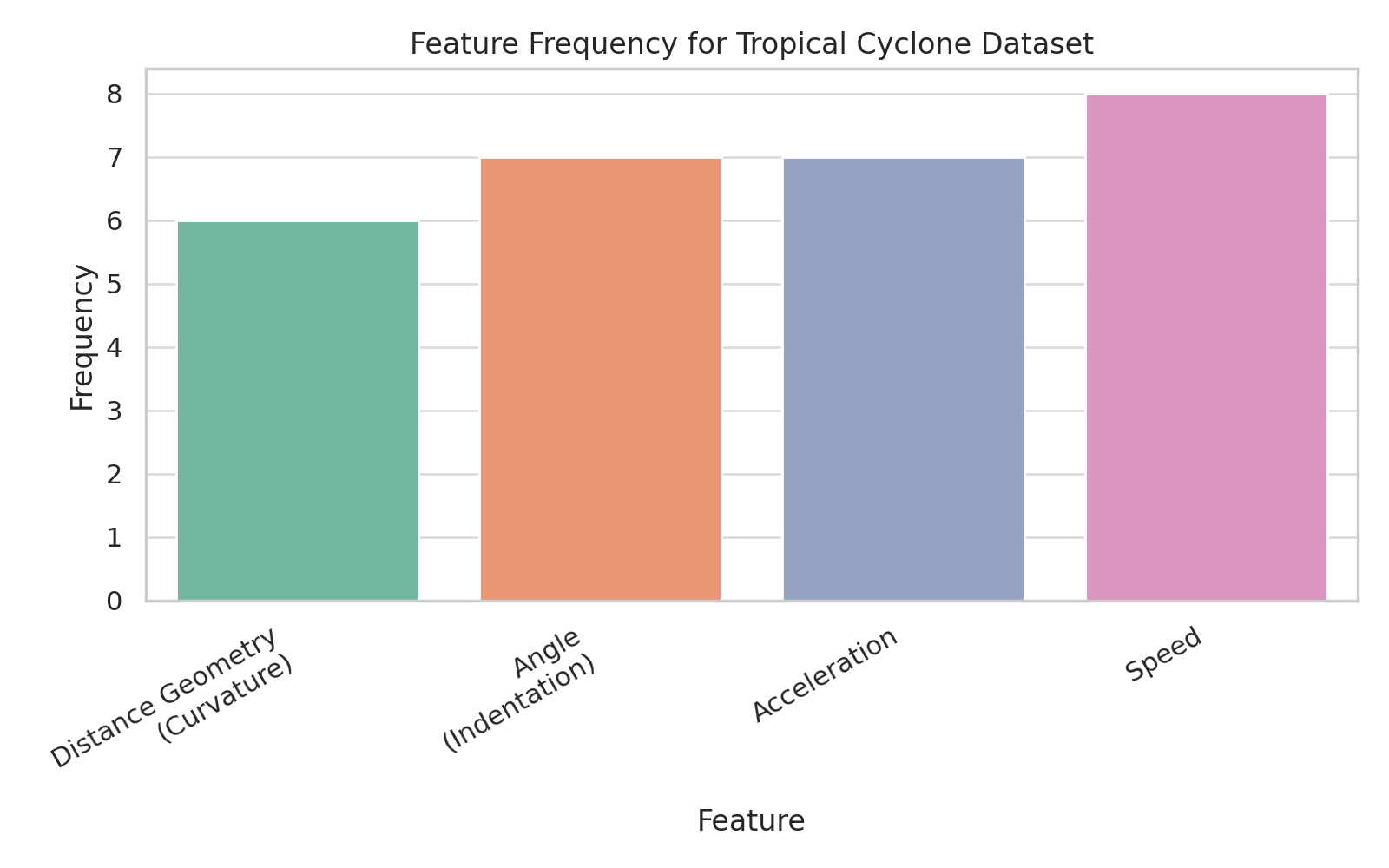}.
    \caption{Frequency of features}
    \label{fig:tc_frequency}
\end{figure}

The results suggests that all categories of the taxonomy was heavily used in the combinations throughout all models. This indicates that the classification problem itself was very complex and required a wide range of features meaning it insinuates cyclone trajectories are hard to classify with a limited feature set. 

\clearpage

\subsection{Cross dataset evaluation summary}

The Tables {\ref{non_tuned_perefences}} and {\ref{tuned_perefences}} summarizes the best-performing feature selection method for each tuned and non-tuned dataset.

\begin{table}[htbp]
\centering
\caption{Preferred Feature Selection Method for Non-tuned Data}
\label{non_tuned_perefences}
\begin{tabular}{|l|l|l|l|}
\hline
\textbf{Model} & \textbf{Arctic Fox} & \textbf{AIS} & \textbf{Tropical Cyclone} \\
\hline
Logistic Regression & Taxonomy-based & Forward Selection & Taxonomy-based \\
Random Forest       & Taxonomy-based & Taxonomy-based & Taxonomy-based \\
XGBoost             & Taxonomy-based & Taxonomy-based & Backward Selection \\
MLP                 & Forward Selection & Taxonomy-based & Backward Selection \\
\hline
\end{tabular}
\end{table}

\begin{table}[htbp]
\centering
\caption{Preferred Feature Selection Method for Tuned Data}
\label{tuned_perefences}
\begin{tabular}{|l|l|l|l|}
\hline
\textbf{Model} & \textbf{Arctic Fox} & \textbf{AIS} & \textbf{Tropical Cyclone} \\
\hline
Logistic Regression & Taxonomy-based & Forward Selection & Taxonomy-based \\
Random Forest       & Taxonomy-based & Forward Selection & Taxonomy-based \\
XGBoost             & Taxonomy-based & Backward Selection & Backward Selection \\
MLP                 & Taxonomy-based & Taxonomy-based & Taxonomy-based \\
\hline
\end{tabular}
\end{table}

When looking at the datasets, the Arctic Fox dataset favors the taxonomy method, suggesting that the taxonomy-based method performs very well with small datasets. This raises the question of whether semantically structured feature sets can perform better when limited sample sets are available. In the context of the AIS dataset, the results are mixed with a 50/50 split, suggesting that it has similar performance in cases where the dataset is easily classifiable. This is further proven by the fact that in the cases where the baseline methods performed better, the taxonomy results were not far behind, as evidenced by Table \ref{tab:AIS_median_f1}. The Tropical Cyclone dataset shows a similar pattern with mixed results. However, none of the results from baseline or the taxonomy-based methods can be considered satisfactory, with the best predictors only having a median Weighted-F1 score of 0.5055.

\subsection{Cross model evaluation summary}

In the case of Logistic regression, taxonomy-based feature selection is preferred, as it accounts for four out of the six best results. The random forest model shows a much stronger inclination to the taxonomy-based method, as seven out of the eight results were from the taxonomy-based method. In contrast with XGBoost, the results do not show any specific preference to a particular approach, with selection being split evenly among baseline and taxonomy-based methods. Random forest models are well regarded for being robust and highly resistant to noise \cite{breiman2001random}. This, coupled with the superior performance of the taxonomy method within the random forest model, suggests that the datasets may contain significant noise and might benefit from enhanced pre-processing or improved sampling strategies.

In the case of MLP, the results were similar to the logistic regression, where six out of the eight results favored the taxonomy-based method. Overall, most of the methods seem to favor the taxonomy-based method over the classical baseline methods, with around $67\%$ of the time, the taxonomy-based method outperforming the baseline. These results per model and dataset seem to mostly confirm that a taxonomy model can derive better results but not in all possible cases \cite{tavakoli2025taxonomical}.

\newpage

\section{Discussion}

\noindent The discussion in this study aims to provide a thorough analysis, with a particular focus on movement trajectory data and feature selection. The primary aim of this study is to evaluate whether the taxonomy-based feature selection approach outperforms the classical forward and backwards selection approaches in the context of high-dimensional trajectory datasets.

\subsection{Key findings}

\vspace{1em}
\textbf{RQ1}: How do existing feature selection approaches perform in the context of high-dimensional trajectory data?

\vspace{1em}

\noindent In the context of classical methods for the fox data set, the best results were gained from the forward selection method in combination with the MLP models, which was a Weighted\_f1 score of 0.5482. As for the hurricane data set the best performer was backwards selection with the combination of XGBoost model with the median Weighted\_f1 score of 0.5055. However, the classical method's performance was more impressive in the context of the AIS data set where it achieved a median Weighted\_f1 score of 0.7597 through the combination of backward selection and the XGBoost model. Overall the performance of the classical methods seems to be lacking except for the AIS dataset. 

\vspace{1em}

\noindent\textbf{RQ2}: How would a taxonomy-based feature selection approach, which systematically organizes the feature set into related categories, perform in the context of trajectory data?

\vspace{1em}

\noindent The taxonomy-based feature selection approach proposed in this study organizes features systematically into two categories: Geometric and Kinematic {\cite{tavakoli2025taxonomical}}, which can be considered a high-level classification. This high level is further classified into subcategories: Curvature, Indentation, Speed, and Acceleration, which can be regarded as low-level taxonomic categories.

The following are some highlights of the results from the taxonomy-based approach. In the Arctic Fox dataset, the taxonomy-based approach was the only feature selection method that generated a median Weighted-F1 score above 0.6000, with values of 0.6080 and 0.6111 for the random forest model and the Boost model, respectively. In the AIS dataset, the taxonomy-based approach performed consistently, getting a median Weighted-F1 score above 0.6700. In contrast, the taxonomy-based approach in the Tropical Cyclone dataset achieved a median Weighted-F1 score above 0.5000 twice, despite the dataset's overall poor performance. Other than that, the random forest model consistently performs better when using the taxonomy-based feature selection approach.

\vspace{1em}

\noindent \textbf{RQ3}: How does the proposed taxonomy-based method compare to conventional feature selection techniques in terms of computational efficiency, performance, and model explainability?

\vspace{1em}

\noindent Tables {\ref{non_tuned_perefences}} and {\ref{tuned_perefences}} show that the taxonomic feature selection method scored higher in 16 out of the total 24 experiments. However, the improvements were not statistically significant. This suggests that, when comparing the performance of the classical and proposed methods, there is no evidence that the taxonomic feature selection consistently outperforms the classical methods in a significant manner.

Interestingly, the taxonomy-based approach consistently outperformed classical approaches in random forest models across all datasets, whereas the taxonomy approach showed competitive results in logistic regression and MLP models. However, in the XGBoost model, both approaches performed equally well. These results were taken into account, including both hyperparameter tuning and excluding it. In the case of the Arctic Fox dataset, the taxonomy-based approach was the only one to exceed the median Weighted-F1 score of 0.6000, indicating its ability to perform well on smaller, varying and complex datasets.

Regarding interpretability, it can be argued that categorizing features into a taxonomic structure enables gaining insights into which categories of the taxonomy the dataset is more sensitive to, may help in facilitating high-level data-driven decision-making although not measured in this study. A frequency analysis was employed to identify the features that were most frequently present in the best-performing taxonomy feature subset for each model. This is only possible due to the high level of information gained through using a taxonomy to organize the features.

Despite the fact that it was not quantitatively measured in this study but taxonomy based methods, speed on selecting features was much faster during the experiments. This is due to the taxonomic structuring of features before selection, whereas classical methods have to take combinations from individual feature levels. Therefore, taxonomy-based feature selection required less computational time and was significantly more efficient, as it reduced the combinatorial space.

Overall the results of this experiment is inconclusive on which feature selection method performs better.

\subsection{Limitations and Methodological Considerations}
Several limitations may have influenced the results:

\vspace{1em}

\textbf{Sample Size and Selection}: The limited and randomly selected sample for the Tropical Cyclone dataset likely increased data variability and noise, thereby lowering the performance of all models. A more systematic sampling strategy, focusing on the last two decades, may yield more consistent and relevant results.

\vspace{1em}

\textbf{Hyperparameter Tuning}: An interesting finding was that hyperparameter tuning sometimes reduced the performance of taxonomy-based selection, possibly due to overfitting or suboptimal parameter ranges. This suggests a need for developing better hyperparameter optimization strategies tailored to taxonomy-based feature selection.

\vspace{1em}

\textbf{Timing and Computational Limitations}: Due to the short period available for this study and hardware limitations, the scope of experiments, hyperparameter tuning, and dataset sizes had to be restricted. As a result, some potentially beneficial configurations and more extensive analyses could not be explored within the available resources.
\newpage

\section{Conclusion and Future Works}

\noindent This thesis proposed a novel feature selection approach based on taxonomies and compares its performance against classical feature selection methods in the context of trajectory data, namely backward and forward selection. 
The proposed taxonomy-based approach systematically organizes the trajectory features into geometric and kinematic features, adding a layer of interpretability to the model's predictive results.

In particular, the taxonomy method consistently outperformed the classical methods for the random forest model, known for its robust and noise-resistant characteristics. 
This outcome suggests that noise is possible within the datasets, and more pre-processing could have improved the results of the taxonomy-based method across all models and datasets. Though not evaluated directly, one observation while experimenting was that it takes much less time to run the taxonomy-related experiment than the classical methods. 
This time reduction suggests that the computational complexity is considerably reduced and needs further research. 
If the decrease in time is considered, the results of the taxonomy-based approach, having comparable or better results, seem much more impressive. Furthermore, an analysis of a dataset's sensitivity to different feature types from the taxonomy was able to gain more insight into the classification problem, which was not possible through the classical methods. 

The results of this study is significant for both academic and practical applications. The taxonomy-based approach's potential opens up more research opportunities for growing work within eXplainableAI (XAI). 
Furthermore, the application of this method could aid a wide variety of practical domains (maritime planning, disaster planning, animal migration and conservation, etc.) in making data-driven decisions with more transparency to the results themselves. 

Despite these contributions, several limitations need to be acknowledged. 
The study was limited to three datasets and four models. 
Next, the taxonomy itself was pre-defined and only contained geometric and kinematic features, where the possibility remains to develop better taxonomies that reflect the characteristics of a specific dataset. 
Additionally, in some cases, the taxonomy-based approach's performance deteriorates. Due to the time limitations, further exploration into better hyperparameters was not possible, and how to effectively configure hyperparameters in the context of taxonomy-based approaches needs to be studied.


\noindent From the results gathered, it is evident that a taxonomy method has high potential within the context of adding explainable AI and reducing computational complexity. Based on this, we propose the following future work. \\

\noindent \textbf{Implementing backward deletion and forward selection.} Here, the idea is to implement two separate selection algorithms within the taxonomy, where a more comprehensive selection approach can be taken. With this approach, there is space to integrate larger taxonomies while decreasing the computational complexity.

\noindent \textbf{Introduction of clustering at a taxonomic level}. Another way to improve the redundancies is to implement a clustering technique to eliminate the redundant features, which can be used to reduce the complexity further without affecting the offered interpretability.

\noindent \textbf{Dynamic taxonomy creation.} In this study, a static taxonomy proposed by Yashar et al. was used. However, the potential for dynamic taxonomy creation while maintaining the taxonomic level interpretation can be explored  
\newpage

%
\newpage

\hypersetup{urlcolor=black}
\bibliographystyle{IEEEtran}
\bibliography{references}
\newpage




\end{document}